\documentclass[10pt,journal,compsoc]{IEEEtran}

\usepackage[dvipsnames]{xcolor}

\ifCLASSOPTIONcompsoc
  \usepackage[nocompress]{cite}
\else
  \usepackage{cite}
\fi

\usepackage{soul}
\usepackage{caption}
\usepackage{subcaption}
\usepackage{graphicx}

\usepackage{hyperref}
\usepackage{amsmath}
\usepackage{amsfonts}
\usepackage{multirow}
\usepackage{caption}
\usepackage{subcaption}
\usepackage{booktabs}
\usepackage{cleveref}
\usepackage{acronym}
\usepackage{todo}
\usepackage{xcolor}

\DeclareMathOperator*{\argmax}{argmax}

\usepackage[dvipsnames]{xcolor}
\newcommand{\C} {black}
\newcommand{\ch}[1] {\textcolor{\C}{#1}}

\begin{document}

\title{COLD Fusion: Calibrated and Ordinal Latent Distribution Fusion for Uncertainty-Aware Multimodal Emotion Recognition}

\author{Mani Kumar Tellamekala, Shahin Amiriparian,  Björn W. Schuller, Elisabeth André, Timo Giesbrecht, Michel Valstar

\IEEEcompsocitemizethanks{

\IEEEcompsocthanksitem Mani Kumar Tellamekala and Michel Valstar are with the Computer Vision Lab, School of Computer Science, University of Nottingham, UK.  \protect\\
E-mail:\{mani.tellamekala, michel.valstar\}@nottingham.ac.uk

\IEEEcompsocthanksitem Shahin Amiriparian and Björn W. Schuller are with the Chair of Embedded Intelligence for Health Care \& Wellbeing, University of Augsburg, Germany. Björn W. Schuller is also with the GLAM – the Group on Language, Audio, \& Music, Imperial College London, UK. \protect\\
Email: \{shahin.amiriparian, schuller\}@uni-a.de

\IEEEcompsocthanksitem Elisabeth André is with the Chair for Human-Centered Artificial Intelligence, University of Augsburg, Germany. \protect\\
E-mail:andre@informatik.uni-augsburg.de

\IEEEcompsocthanksitem Timo Giesbrecht is with Unilever R\&D Port Sunlight, UK. \protect\\
Email: timo.giesbrecht@unilever.com
}
\thanks{}}

\markboth{ACCEPTED TO IEEE TRANSACTIONS ON PATTERN ANALYSIS AND MACHINE INTELLIGENCE}
{}

\IEEEtitleabstractindextext{%
\begin{abstract}
Automatically recognising apparent emotions from face and voice is hard, in part because of various sources of uncertainty, including in the input data and the labels used in a machine learning framework. This paper introduces an uncertainty-aware multimodal fusion approach that quantifies modality-wise aleatoric or data uncertainty towards emotion prediction. We propose a novel fusion framework, in which latent distributions over unimodal temporal context are learned by constraining their variance. These variance constraints, Calibration and Ordinal Ranking, are designed such that the variance estimated for a modality can represent how informative the temporal context of that modality is w.r.t. emotion recognition. When well-calibrated, modality-wise uncertainty scores indicate how much their corresponding predictions are likely to differ from the ground truth labels. Well-ranked uncertainty scores allow the ordinal ranking of different frames across different modalities. To jointly impose both these constraints, we propose a softmax distributional matching loss. Our evaluation on AVEC 2019 CES, CMU-MOSEI, and IEMOCAP datasets shows that the proposed multimodal fusion method not only improves the generalisation performance of emotion recognition models and their predictive uncertainty estimates, but also makes the models robust to novel noise patterns encountered at test time.
\end{abstract}

}

\maketitle

\IEEEdisplaynontitleabstractindextext

\IEEEpeerreviewmaketitle

\IEEEpeerreviewmaketitle

\section{Introduction}
\label{sec:intro}

\begin{figure}
    \centering
    \includegraphics[width=0.99\linewidth]{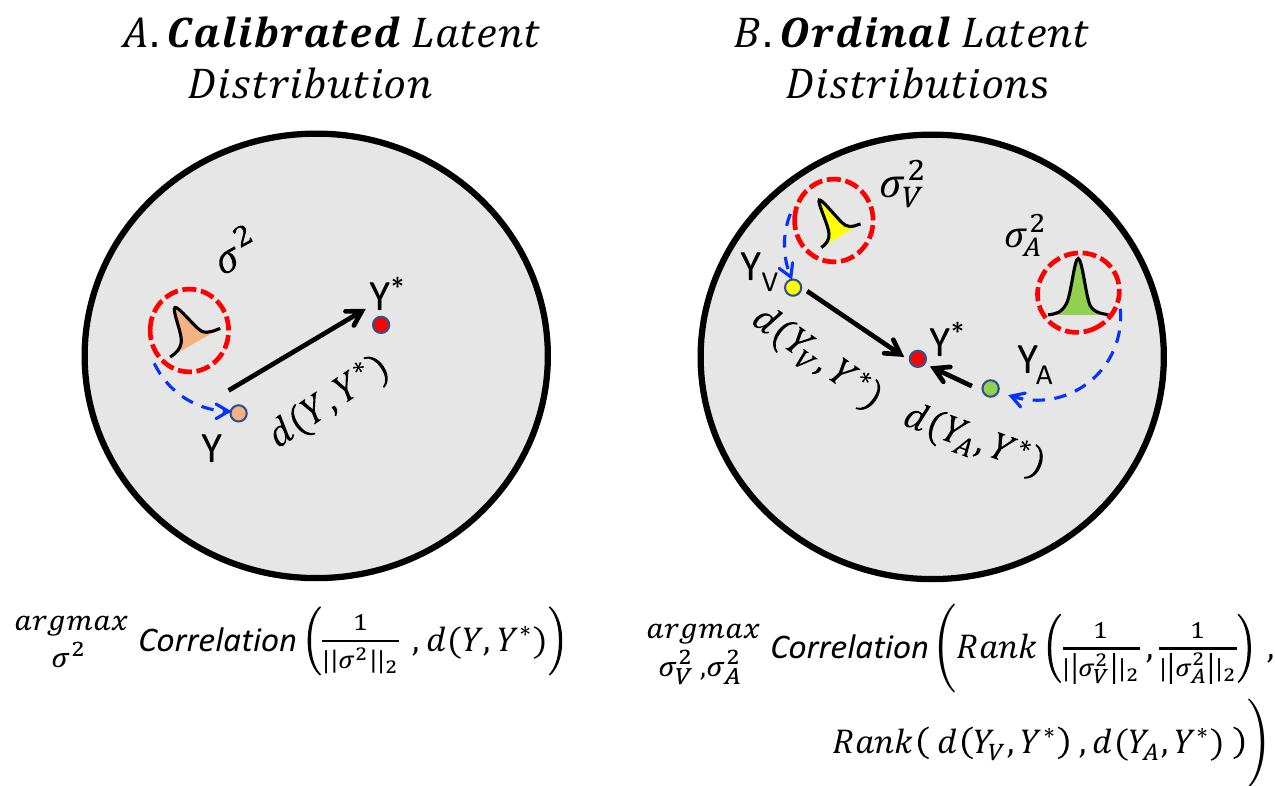}
    \caption{The proposed latent distribution learning approach for multimodal fusion ($Y_V$ and $Y_A$ -- unimodal predictions, $Y^*$ -- target label, and $d$ -- a distance function): The latent distributions' variance values are learned to represent how informative the temporal context of each modality is, by applying the constraints: \textit{\textbf{A. Calibrated Latent Distribution:}} Temporal context is modelled by a latent distribution learned under the \textit{calibration constraint} so that its variance can act as a proxy for the target prediction error ($d(Y, Y^*$). \textit{\textbf{B. Ordinal Latent Distributions:}} The variance values of audio and visual temporal context distributions ($\sigma^2_V$ and $\sigma^2_A$) are learned under the \textit{ordinal constraint} so that the audio and visual modalities are ranked according to their prediction errors ($d(Y_V, Y^*)$ and $d(Y_A, Y^*)$).}
    \label{fig:trailer_fig}
\end{figure}

\begin{figure*}
    \centering
    \includegraphics[width=0.90\linewidth]{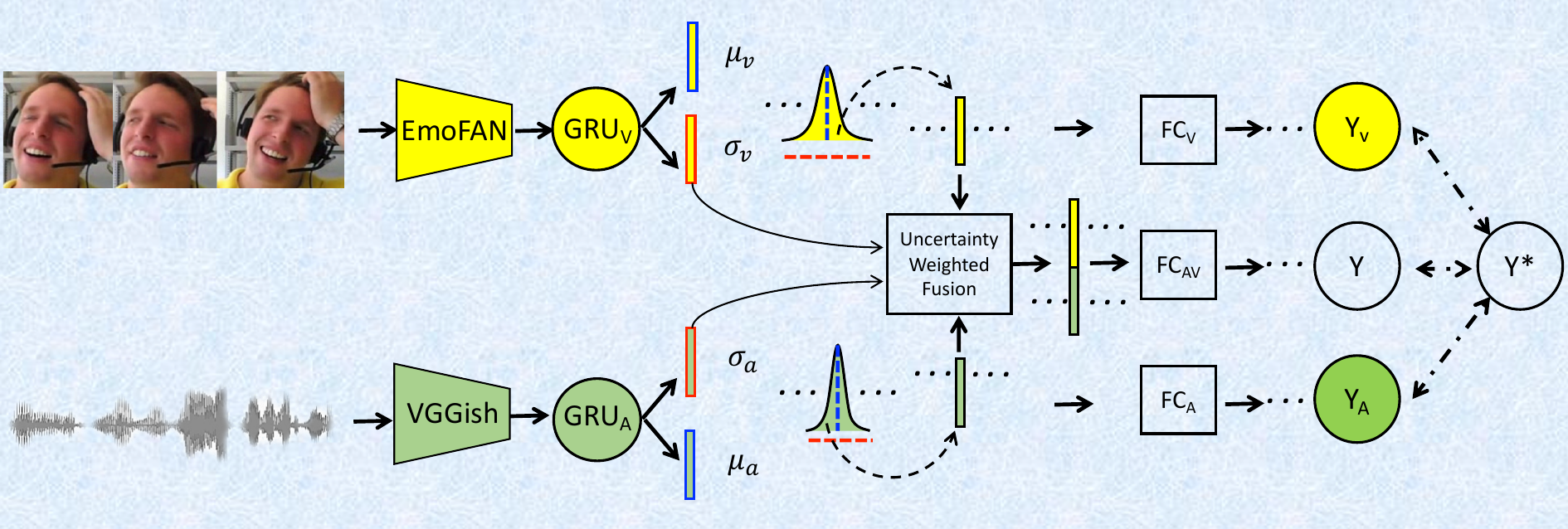}
    \caption{Overview of the proposed approach to an uncertainty-aware audiovisual fusion for emotion recognition: Modelling latent distributions over unimodal temporal context vectors to derive modality-wise uncertainty guided fusion weights. A detailed description of our approach is given in~\Cref{sec:fusion_background_notations}.}
    \label{fig:cold_main}
\end{figure*}

\IEEEPARstart{L}{earning} to fuse task-specific information from multiple modalities is a fundamental problem in Machine Learning. At its core, this problem  entails estimating how informative each modality is towards predicting the labels of a target task. For example, consider the task of automatically recognising emotional expressions from a video in which a person is talking with a face mask covering. In such a scenario, for effectively fusing information from the audio and visual modalities, the model must be aware of how informative the facial and vocal streams are w.r.t the target task separately. Thus, modality-wise uncertainty-aware fusion is a natural approach to multimodal learning. 

In this work, we formulate an uncertainty-aware fusion method for the task of apparent emotion recognition from multimodal inputs. The proposed multimodal fusion framework is based on probabilistic modelling of unimodal temporal context related to emotional expressions. This probabilistic temporal modelling approach aims to capture the richness of the temporal context in terms of emotional expressions present in a given modality, and use that information in deciding the degree of importance of each modality towards recognising apparent emotions.

In the proposed method, we first estimate the uncertainty of  unimodal temporal inputs, and then apply those uncertainty estimates in computing modality-wise fusion weights. \textit{In particular, we aim to estimate the aleatoric component of uncertainty~\cite{hullermeier2021aleatoric} associated with different modalities for improved emotion recognition performance.} Unlike the epistemic component of uncertainty, which can be explained away with more data,  aleatoric uncertainty captures noise or stochasticity that is inherent to an input signal. To give an example, in recognising emotional expressions from face images, the epistemic component can describe the uncertainty due to insufficient data for 'happy' class whereas the aleatoric component captures the uncertainty caused by factors like occluded facial regions, low-resolution face images, etc. In this work, we focus on estimating modality-wise aleatoric uncertainty in a multimodal emotion recognition model.    

Being an intrinsically temporal and multimodal phenomenon, emotion recognition from multimodal inputs is a long-standing challenge in Affective Computing~\cite{schuller2012avec, valstar2013avec, ringeval2019avec}. A meta-analysis presented in~\cite{d2015review} has shown that although emotion recognition can benefit from multimodal fusion in general, performance improvements are not significant when it comes to spontaneous emotions. We believe that uncertainty-aware multimodal fusion may have the potential to address this challenge, considering that the intensity of spontaneous emotions embedded in different modalities are likely to vary dynamically over time~\cite{nicolaou2011continuous,zeng2008survey}. 

Although Deep Neural Networks (DNNs) have been extensively applied to multimodal emotion recognition~\cite{rouast2019deep,noroozi2017audio,schoneveld2021leveraging,gerczuk2021emonet}, estimating modality-wise uncertainty for improved fusion performance is a relatively  unexplored avenue. However, modelling predictive uncertainty (or confidence, its opposite) in DNNs received widespread attention in recent years~\cite{laurent2023packed,guo2017calibration,mukhoti2020calibrating}, motivated by the observation that DNNs tend to make over-confident predictions~\cite{nguyen2015deep,szegedy2013intriguing}. Most existing efforts towards uncertainty or confidence estimation in DNNs~\cite{guo2017calibration,kumar2018trainable} focus solely on reducing miscalibration errors, i.e., the mismatch between expected model estimation errors and their corresponding confidence scores. Recently, as an alternative perspective, Moon et al.~\cite{moon2020confidence} introduced the idea of learning to rank confidence scores for identifying the most reliable predictions.

In this work we argue that the estimated uncertainty scores must be simultaneously both \textit{well-calibrated}  and \textit{well-ranked (ordinal)}. The former is needed to accurately represent the correctness likelihood of a prediction for \textbf{an individual sample}. The latter is essential to effectively order predictions for \textbf{a group of samples} according to their correctness likelihoods. In other words, if an uncertainty estimate of an individual sample is well-calibrated, in the absence of its ground truth, the uncertainty score can serve as a proxy for its expected prediction error. If the uncertainty scores associated with different predictions are well-ranked or maintain ordinality, then one can use them to order their corresponding samples in terms of their reliability towards the target prediction, and to distinguish the most informative samples from the least informative samples.

For multimodal temporal learning, it is critical to estimate how informative the predictions made for different frames in different unimodal sequences are, towards estimating a common target label, so that the target-specific information can be reliably integrated~\cite{yang2017deep}. In this work, we hypothesise that jointly learning these two properties -- calibration and ordinality -- can lead to more reliable uncertainty estimates for each modality, facilitating more effective uncertainty-weighted temporal context fusion. Based on this hypothesis, we propose an uncertainty modelling method that imposes the calibration and ordinality constraints jointly, as~\Cref{fig:trailer_fig} illustrates.  

For example, consider the task of classifying whether a person's apparent emotional state as either 'happy' or 'neutral' by analysing a face image sequence and its speech signal. Assume that the face is covered with a mask in most frames, making the face modality less informative than the speech modality. In a unimodal setting, the face and speech classifiers are trained separately to output their corresponding 'happy' class probabilities. When well calibrated, these output probabilities should reflect the true correctness likelihoods of face and speech models' predictions. Similarly, when constrained by ordinal ranking, the speech model's output probability must be higher than the face model's probability, reflecting the relative uncertainty levels of the face and speech modalities w.r.t. each other.

In this work, we condition the unimodal latent distributions' variance vectors such that they represent the information different modalities contain w.r.t.\ predicting emotion. The proposed method can be viewed as an uncertainty-aware extension of classical late fusion, but here the fusion is applied in the latent space of unimodal temporal context embeddings. This approach is different from a simple confidence-weighted late fusion model in which uncertainty is modelled directly over the unimodal output predictions.

In our proposed framework, denoted as Calibrated Ordinal Latent Distributions (COLD), we first learn the latent distributions (multivariate normal distributions) over the temporal context of audio and visual modalities separately, as~\Cref{fig:cold_main} shows. We model the variance values of the audio and visual latent distributions, $\sigma_V$ and $\sigma_A$, as the confidence measures towards emotion prediction. We design a novel training objective based on softmax distributional matching to encourage the  variance norm values in each modality to be: (a) strongly correlated with the correctness likelihood of the unimodal predictions, and (b) ordinal in nature to effectively rank the relevance of different modalities towards emotion recognition. Thus, the calibrated and ordinal unimodal variance scores are learnt for effective uncertainty-weighted fusion, as shown in~\Cref{fig:cold_main}. 

We evaluate the proposed COLD fusion approach on: (a) dimensional emotion recognition from face and voice modalities in the AVEC 2019 CES~\cite{ringeval2019avec} and IEMOCAP~\cite{busso2008iemocap} datasets, and (b) categorical emotion recognition from face, voice and text modalities in the CMU-MOSEI~\cite{zadeh2018multimodal} and IEMOCAP datasets. Compared to the uncertainty-unaware fusion baselines, COLD fusion demonstrates noticeably better results on different multimodal emotion recognition tasks evaluated in this work. For example, in dimensional emotion regression tasks COLD fusion shows $\sim$6\% average relative improvement over the best performing fusion baseline. Similarly, in the case of categorical emotion classification COLD fusion achieves $\sim$8.2\% relative accuracy improvement over the existing state-of-the-art model. Furthermore, we assess the robustness of different fusion models at test time by inducing noise into the visual modality through face masking. With the faces masked in 50\% of the evaluation sequences, COLD fusion achieves $\sim$17\% average relative improvement over the best fusion baseline.

The key contributions of our work are as follows:

\begin{itemize}
    \item We propose an uncertainty-aware multimodal fusion method that dynamically estimates the fusion weights to be assigned to unimodal features.
    \item We demonstrate how to jointly learn \textit{well-calibrated} and \textit{well-ranked} unimodal uncertainty estimates. For this purpose, we propose a simple softmax distributional matching loss function that applies to both regression and classification models. 
    \item On \ch{both dimensional and categorical emotion recognition tasks}, the proposed fusion approach shows noticeable performance gains and improved robustness to novel noise patterns encountered at test time. 
\end{itemize}

\section{Related Work}
\label{sec:rel_work}

\noindent \textbf{Multimodal Affect Recognition.} Humans rely primarily on visual (faces) and audio (voices) modalities to encode and express their affective or emotional states. Recognising dimensional emotions, valence (how pleasant an emotion is) and arousal (how active an emotion is), and categorical emotions (happy, sad, disgust, etc) from multiple modalities, is a widely studied problem in various prior works~\cite{zeng2008survey,gunes2011emotion}, ranging from the almost a decade-long running annual AVEC challenge series~\cite{schuller2012avec, valstar2013avec, ringeval2019avec} to the recently introduced MuSe challenge~\cite{stappen2020muse,stappen2021muse,christ2022muse} and ABAW challenge~\cite{kollias2020analysing,kollias2021analysing}. Beyond audiovisual modalities, some recent works (e.g.~\cite{mittal2020emoticon,yang2022emotion}) explored how to fuse a wide range of contextual cues for reliably recognising expressed emotions.

We refer the reader to Poria et al.~\cite{poria2017review}, Roust et al.~\cite{rouast2019deep}, Jiang et al.~\cite{jiang2020snapshot} and Zhao et al.~\cite{zhao2021emotion} for comprehensive surveys of affect recognition in multimodal settings and contemporary deep learning-specific advancements in it. Since our main focus in this work is on uncertainty-aware fusion models for emotion recognition, we review the literature closely related to the following key research topics: i) uncertainty modelling for emotion and expression recognition, ii) uncertainty-aware multimodal fusion, iii) calibrated uncertainty, and iv) ranking-based uncertainty. 

\noindent \textbf{Uncertainty Modelling for Emotion and Expression Recognition.} In categorical facial expression recognition tasks, modelling predictive uncertainty is studied in several recent works~\cite{she2021dive,zhang2021relative,wang2020suppressing}, by estimating uncertainty in the space of low-dimensional feature embedding outputs from a Convolutional Neural Network (CNN) backbone. On the other hand, directly predicting emotion label uncertainty is explored in~\cite{foteinopoulou2021estimating}, but only in unimodal (video-only) settings. For uncertainty-aware multimodal emotion recognition, some prior works applied Kernel Entropy Component Analysis (KECA)~\cite{zeng2005multi} and Multi Modal-Hidden Markov Models (MM-HMMs)~\cite{xie2013multimodal} by predicting modality-specific uncertainty measures for estimating the fusion weights.  

Noting the limitations of deterministic function learning in DNNs for uncertainty modelling, Dang et al.~\cite{dang2017investigating} explored the application of Gaussian Process (GP) Regression to the fusion of emotion predictions. With the same motivation, in Affective Processes (APs)~\cite{sanchez2021affective,mani2021stochastic}, Neural Processes~\cite{garnelo2018conditional,garnelo2018neural} have been applied to the task of emotion recognition. By combining the abilities of GPs to learn function distributions with DNN's representation learning abilities, APs demonstrated superior generalisation performance over deterministic function learning models. Building on this idea of stochastic modelling of temporal functions, recently APs have been extended to multimodal settings in~\cite{tellamekala2022modelling} based on a strictly model-based fusion approach, demonstrating impressive emotion recognition results. However, for uncertainty-aware temporal context modelling, APs heavily rely on the proxy labels predicted by a separate pre-trained backbone and a complex encoder-decoder formulation. In contrast to APs, our method aims to model the temporal context uncertainty in a model-agnostic fashion, by just altering the output head of simple CNN+RNN models that are trained using some novel constrained optimisation objectives.

All the aforementioned methods demonstrated the potential of uncertainty-aware emotion recognition models over their uncertainty-unaware counterparts in general. However, they ignore two important aspects of uncertainty modelling: calibration and ordinality (ranking). In this work, we aim to demonstrate the significance of these two properties by hypothesising that learning well-calibrated and well-ranked uncertainty estimates is critical for improving multimodal emotion recognition performance. 

\noindent \textbf{Uncertainty-Aware Multimodal Fusion.}  For multimodal sensor fusion, several prior works~\cite{zeng2005multi,schorgendorfer2006extended,grosse2008confidence,papandreou2009adaptive} explored uncertainty-aware or confidence-weighted averaging techniques for classic machine learning models before the advent of Deep Neural Networks (DNNs). Recently, Subedar et al.~\cite{subedar2019uncertainty} applied Bayesian DNNs for uncertainty-aware audiovisual fusion to improve human activity recognition performance. Similarly, Tian et al.~\cite{tian2020uno} explored the use of uncertainty estimation in fusing the softmax scores predicted using CNNs for semantic segmentation. \ch{Other notable approaches to uncertainty-aware multimodal fusion are based on optimal transport for cross-modal correspondence~\cite{pramanick2022multimodal}, random prior functions~\cite{wang2022uncertainty}, boosted ensembles~\cite{sarawgi2021uncertainty}, and factorised deep markov models~\cite{zhi2020factorized}.} 

Although all the aforementioned methods demonstrated critical advantages over the models that predict only point estimates, they do not study the calibration properties of the estimated uncertainty scores. Further, such DNN models focus mainly on modelling absolute uncertainty estimates, whereas our focus is on jointly \textbf{learning the calibrated and relational uncertainty estimates} in an end-to-end fashion introducing a novel softmax distributional matching loss.

\noindent \textbf{Calibrated Uncertainty.} As DNNs tend to make overconfident predictions~\cite{nguyen2015deep,szegedy2013intriguing}, confidence calibration has received significant attention in recent years~\cite{nguyen2015deep,szegedy2013intriguing}. Calibrating confidence or uncertainty estimates involves maximising the correlation between predictive accuracy values and predictive uncertainty scores. A wide variety of calibration techniques, particularly in classification settings, can be broadly categorised into explicit and implicit calibration categories~\cite{wang2021rethinking}.  In the former category, two types of post-hoc methods, binning-based and temperature-scaling, are applied to increase the reliability of DNN confidence estimates~\cite{guo2017calibration,minderer2021revisiting}. In binning-based methods such as non-parametric histogram binning~\cite{zadrozny2001obtaining}, calibrated confidence is estimated based on the average count of positive-class instances in each bin. This method is extended to jointly optimise the bin boundaries and their predictions in Isotonic Regression~\cite{zadrozny2002transforming}. Temperature-scaling methods can be viewed as generalised versions of Platt scaling~\cite{platt1999probabilistic} using logistic regression for calibrating the class probabilities. We use temperature-scaling as a calibration baseline~\cite{hinton2015distilling,guo2017calibration} to compare against the uncertainty calibration performance of the proposed method, due to its simplicity. 

Implicit calibration methods tailor the training objective of DNNs to minimise the prediction error and calibration error simultaneously. Addressing the limitations of standard cross-entropy loss w.r.t.\ confidence calibration, various alternative loss functions such as focal loss~\cite{mukhoti2020calibrating}, maximum mean calibration error~\cite{kumar2018trainable}, and accuracy vs uncertainty calibration~\cite{krishnan2020improving}, have been investigated recently. Calibrating regression models is relatively under-explored compared to the classification. Some recent works~\cite{kuleshov2018accurate,song2019distribution,utpala2020quantile} made attempts to extend some of the aforementioned calibration techniques to continuous-valued predictions. 

\noindent \textbf{Ordinal or Ranking-based Uncertainty.} In the existing uncertainty modelling works, the ordinal property of uncertainty estimates received less attention compared to the calibration property, which partly motivated the method introduced in this paper.  Li et al.~\cite{li2021learning} proposed to model data uncertainty by inducing ordinality into probabilistic embeddings of face images. Towards uncertainty-aware regression problems, the results reported in~\cite{li2021learning} highlighted the key limitations of deterministic unordered embeddings compared to the probabilistic ordinal embeddings. Although not strictly ordinal, relative uncertainty modelling is explored for facial expression recognition in~\cite{zhang2021relative}.  

Other closely related works approached the problem of ordinal ranking of uncertainty estimates with different objectives such as failure prediction~\cite{corbiere2019addressing}, out-of-distribution detection~\cite{roady2019out}, and selective classification~\cite{geifman2017selective}. Fundamentally, all these objectives necessitate a method that can train the model to output well-ranked confidence or uncertainty scores. Among these existing methods, the one most closely related  to ours is by Moon et al.~\cite{moon2020confidence}, which proposes a Correctness Ranking Loss (CRL). CRL  directly imposes ordinal ranking constraints on the confidence estimates of a DNN classifier. Similar to CRL, our proposed softmax distributional matching loss also constrains the ordinal-ranking property of uncertainty estimates. However, in addition to ordinal ranking, our method imposes the calibration property as well, most importantly by controlling the latent distribution variance, unlike in CRL. Moreover, our formulation generalises the idea of calibrated and ordinal uncertainty estimates to both classification and regression settings, using a common loss function computation.

\section{Model-Agnostic Fusion Baselines}
\label{sec:fusion_background_notations}

Before introducing our uncertainty-aware multimodal fusion formulation, we briefly discuss the general multimodal fusion techniques w.r.t.\ audiovisual emotion recognition and introduce the related notations. A fundamental question in multimodal learning concerns the optimal stage to perform fusion~\cite{baltruvsaitis2018multimodal}. We consider the following three typical model-agnostic fusion methods as the baselines: feature fusion, temporal context fusion, and prediction fusion.

\noindent \textbf{Preliminaries and Notations.} As~\Cref{fig:cold_main} illustrates, given a face video clip $X_V$ with $N$ frames and its corresponding speech signal $X_A$, using overlapping time windows, we first create $N$ speech segments that correspond to the $N$ visual frames. Here, we assume that both the signals $X_V$ and $X_A$ are annotated with a common dimensional emotion label, $Y^*=[Y^*_{valence}, Y^*_{arousal}]$ (either per-frame or per-sequence). We extract sequences of per-frame low dimensional features $(Z_V, Z_A)$ from the face video and speech inputs using a two-stream network. This network is composed of a 2D CNN $f_V$ and a 1D CNN $f_A$ for processing the face images and speech segments respectively, $f_V: X_V \rightarrow [z_V^1, z_V^2, ..., z_V^N]$ and $f_A: X_A \rightarrow [z_A^1, z_A^2, ..., z_A^N]$. For unimodal emotion recognition, we process the temporal context from each modality separately from $Z_V$ and $Z_A$ using different temporal networks $g_V: Z_V \rightarrow Y_V$ and $g_A: Z_A \rightarrow Y_A$ to predict the emotion labels $Y_V$ and $Y_A$.

\noindent \textbf{Feature Fusion} or early fusion integrates frame-level emotion cues present in the audiovisual features $Z_V$ and $Z_A$ (e.g.,~\cite{zhang2017learning}), not accounting for commonly encountered temporal misalignment between different modalities~\cite{lingenfelser2016asynchronous}. Here, we concatenate the per-frame audiovisual features into a single sequence, $Z=[Z_V, Z_A]$, then pass it to a common temporal network $g_{AV}: Z \rightarrow Y$ to predict emotion labels.

\noindent \textbf{Decision Fusion} combines the unimodal emotion predictions $Y_V$ and $Y_A$ (e.g.,~\cite{ringeval2015prediction}). Here, we apply predictive confidence based weighted averaging to perform the late fusion. Unlike early fusion, late fusion does not leverage the low-level correspondences among the emotion cues distributed over the audio and visual streams~\cite{baltruvsaitis2018multimodal}. 

\noindent \textbf{Temporal Context Fusion} or simply context fusion integrates sequence-level emotion information aggregated in the form of audiovisual temporal context vectors $h_V^i$ and $h_A^i$ for frame $i$, produced by the temporal networks $g_V$ and $g_A$ respectively. This method is also referred to as `feature fusion with RNNs' or `mid-level' fusion in some prior works~\cite{rouast2019deep,tzirakis2017end}. Note that here temporal context or simply context at $i^{th}$ frame refers to the emotion information present in frame $i$ w.r.t.\ the emotion information carried by remaining frames in the input sequence. As a result, unlike early fusion, context fusion is bound to suffer less from the temporal misalignment between the emotion-related semantics of audio and visual feature sequences. Further, context fusion benefits from the low-level audiovisual correspondences in the emotion space, in contrast to late fusion.

Considering the above-mentioned critical advantages of temporal context fusion, in this work, we propose to learn an uncertainty-aware context fusion model for multimodal emotion recognition as discussed below. 

\section{Proposed Method}
\label{sec:method}

\begin{figure*}
    \centering
    \includegraphics[width=0.9\linewidth]{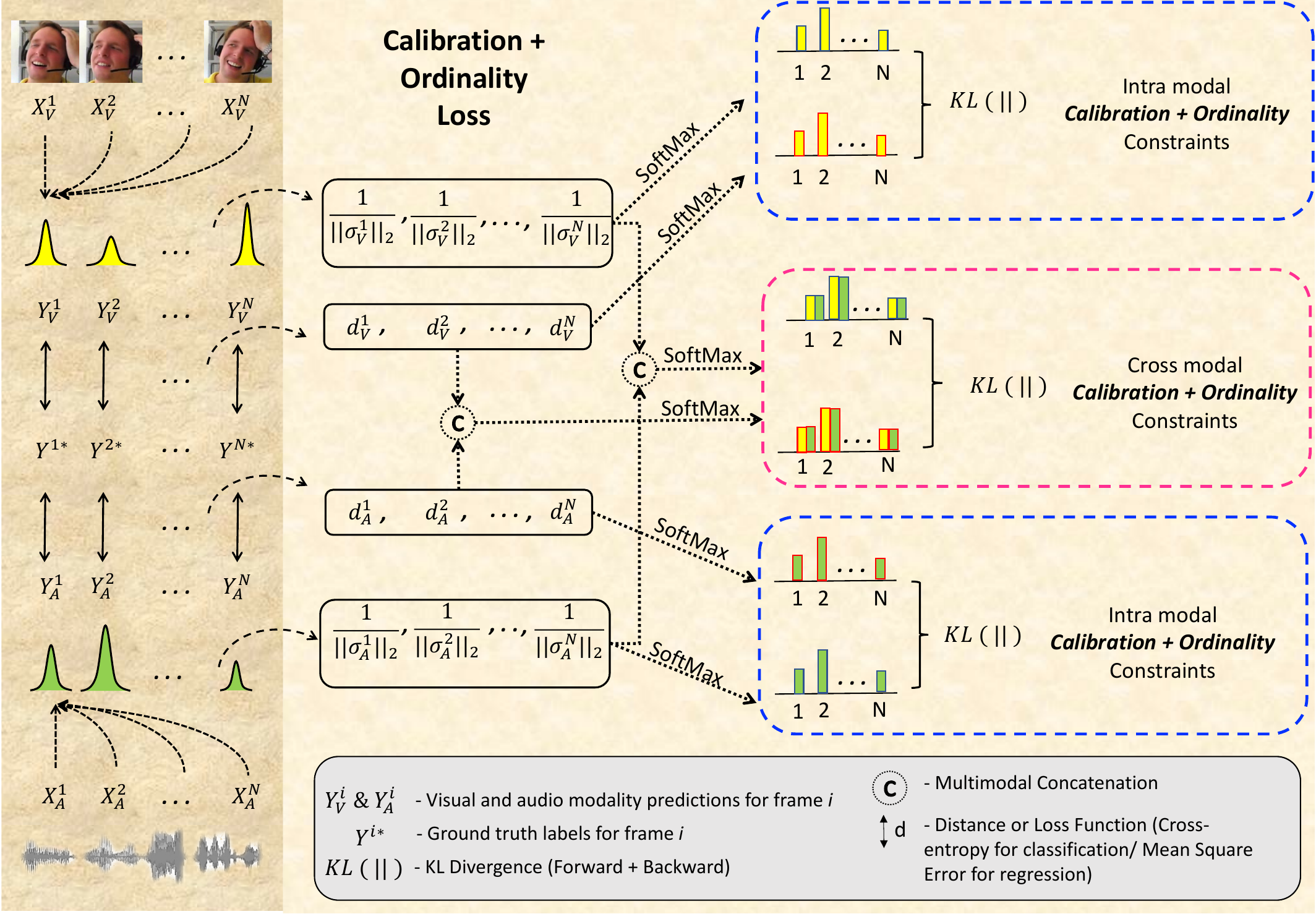}
    \caption{COLD fusion training loss computation: To simultaneously impose the calibration and ordinality constraints on the unimodal latent distributions' variance vectors, we minimise the softmax distributional matching loss (KL divergence)  between the distance vectors $[d^i]$ and variance-norm vectors $[\frac{1}{ \| {\sigma^i}^2 \|_{2} }]$, in both intramodal and crossmodal settings.}
    \label{fig:cold_loss}
\end{figure*}

\Cref{fig:cold_loss} illustrates our proposed solution to uncertainty-aware multimodal fusion. Although this section describes the proposed fusion only in audiovisual settings, note that it can be easily extended to tasks with more than two modalities. In this section, we first discuss how we estimate modality-wise uncertainty by learning unimodal latent distributions over the temporal context, and we present our approach to derive the fusion weights based on unimodal context variance. Then, we introduce two key optimisation constraints imposed on the variance norms of unimodal latent distributions and describe their implementations.

\subsection{Uncertainty-Aware Audiovisual Context Fusion}
Quantifying modality-wise uncertainty towards predicting a common target label is crucial to improve multimodal fusion performance. Our objective is to first quantify intramodal uncertainty in the temporal context space, and then use the estimated uncertainty scores to derive the fusion weights. To this end, we propose to learn unimodal latent distributions over the temporal context of the audio and visual modalities separately, as discussed below.

\subsubsection{Latent Distributions over Unimodal Temporal Context} \Cref{fig:cold_main} illustrates how we modify the temporal networks (Gated Recurrent Unit(GRU)-RNNs) $g_V$ and $g_A$ to output the parameters (mean and variance) of multivariate normal distributions $\mathcal{N}(\mu_V^i,\, {\sigma^i_V}^2)$ and $\mathcal{N}(\mu_A^i,\, {\sigma^i_A}^2)$ over the audio and visual temporal context vectors, respectively. Here, the term `temporal context' refers to the hidden state outputs from the corresponding unimodal GRU blocks ($g_A$ or $g_V$). For each modality separately, we learn this hidden state output as a multivariate normal distribution, instead of a typical deterministic embedding vector. We presume that these unimodal latent distributions are capable of representing modality-wise emotion information more effectively than deterministic embeddings. 

Given a sequence of frames, [$X_1$, $X_2$, ..., $X_T$], in order to predict their corresponding target variables [$Y^*_1$, $Y^*_2$, ..., $Y^*_T$] it is important to learn the underlying temporal context information, which is a function of the frames present in the input sequence as well as the order in which they appear. By modelling the temporal context as a probability distribution, we propose to use the prediction error $\| Y_i - Y^*_i \|_{2}$ to constrain the contribution of each frame $X_i$ in terms of its explained variance of the overall temporal context. Here, the idea of frame-wise explained variance of the temporal context refers to how much information a particular frame holds given all the rest of the frames, towards predicting the target variable $Y^*_i$. Thus, the higher the explained variance of a particular frame $X_i$, the more informative it is for accurately predicting the target variable. 

Here our aim is to first estimate the informativeness of each modality towards the task of recognising emotions. To this end, we learn the temporal context variance such that it may represent how informative the temporal context of a particular modality is. For example, consider an audio-visual sequence in which all the audio frames have the same emotion (e.g. neutral tone), whereas the visual frames have more variations in terms of the emotional expressions. In this case, the fusion model must give more importance to the visual frames compared to the audio frames when predicting emotions. Guided by this intuition, our formulation aims to capture the emotion-related variance in the temporal context of each modality separately.

It is important to note the difference between the absolute variance of the temporal context distribution learned from all the frames and the explained temporal context variance of an individual frame. While the former can be thought of as a proxy metric for uncertainty measurement, the latter can be viewed as a per-frame information metric w.r.t the target prediction. For the sake of simplicity, throughout this work, we use the term 'context variance' in order to refer to the explained variance of temporal context for a given frame in an input sequence. The above argument can be extended to a multimodal fusion setting as well, in which the explained temporal context variance of a particular modality can be used as a proxy for how informative that modality is w.r.t predicting a common target variable.

We model the variance of a unimodal latent distribution as a proxy for how informative that modality is w.r.t.\ predicting the target emotion, and we use the inverse of variance values to quantify how uncertain a particular modality is towards predicting emotion labels. Note that the potential of signal variance-based uncertainty modelling for multimodal fusion was already demonstrated in~\cite{evangelopoulos2013multimodal}. Similarly, learning latent distribution variance was determined to be capable of uncertainty modelling in~\cite{sanchez2021affective}. Inspired by these ideas, we model the unimodal context variance norm values $\|\sigma^2_V\|_{2}$ and $\|\sigma^2_A\|_{2}$ to estimate how certain the audio and visual modalities are about predicting the emotion labels. Our approach to derive variance-based fusion weights for integrating the audiovisual information is discussed below. 

\subsubsection{Context Distribution Variance-Based Fusion Weights}
\ch{For an input frame with index $i$, given its unimodal latent distributions $\mathcal{N}(\mu_V^i,\, {\sigma^i_V}^2)$ and $\mathcal{N}(\mu_A^i,\,  {\sigma^i_A}^2)$ over its visual and audio temporal context embeddings separately}, we first compute the $L2$ norms of their variance values $\| {\sigma^i_V}^{2} \|_{2}$ and $\| {\sigma^i_A}^2 \|_{2}$. As discussed above, these variance norm values are assumed to represent modality-specific certainty or informativeness w.r.t.\ predicting the target emotions. By normalising the variance norm values of the audio and visual modalities, we derive fusion weights that are used in a simple linear fusion model of the audiovisual temporal context ($h^i_{VA}$) : 

\begin{equation}
h^i_{VA} = w^i_V * h^i_V + w^i_A * h^i_A,
    \label{eq:fusion_model}
\end{equation}

\noindent where $h^i_V$ and $h^i_A$ denote the visual and audio temporal context vectors, and $w^i_V$ and $w^i_A$ denote their corresponding weight values. The temporal context vectors $h^i_V$ and $h^i_A$ are sampled from their respective latent distributions, $h^i_V\sim\mathcal{N}(\mu_V^i,\, {\sigma^i_V}^2)$ and $h^i_A\sim\mathcal{N}(\mu_A^i,\, {\sigma^i_A}^2)$ during training. At test time, we set $h^i_V$ and $h^i_A$ to their corresponding mean vectors $\mu_V^i$ and $\mu_A^i$ for evaluation purposes. 

Based on the unimodal context variance norm values ($\| {\sigma^i_V}^{2} \|_{2}$ and $\| {\sigma^i_A}^2 \|_{2}$), the weight values $w^i_V$ and $w^i_A$ in~\Cref{eq:fusion_model} are computed as:

\begin{equation}
    w^i_V = \dfrac{\| {\sigma^i_V}^{2} \|_{2} }{(\| {\sigma^i_V}^{2} \|_{2} + \| {\sigma^i_A}^{2} \|_{2} )}   ,
    w^i_A = \dfrac{\| {\sigma^i_A}^2 \|_{2} }{( \| {\sigma^i_V}^2 \|_{2} + \| {\sigma^i_A}^2 \|_{2} )}.
    \label{eq:fusion_weights}
\end{equation}

Context variance modelling seems to be a simple yet effective approach to uncertainty-aware audiovisual fusion, yet  learning audiovisual latent distributions with well-conditioned variance ranges is non-trivial in practice, as we show later in the experiments. To condition the variance values that can effectively capture intramodal uncertainty w.r.t.\ predicting the target labels, we define a more principled model training that applies two key optimisation constraints: Calibration and Ordinality.

\subsection{COLD: Calibrated and Ordinal Latent Distributions}
To effectively learn the unimodal latent distributions for uncertainty-aware fusion, we propose to condition their variance values by applying optimisation constraints to the model training objective. We achieve this conditioning by imposing two key constraints: Calibration and Ordinality (or ranking) on the latent distribution variance vectors. When well-calibrated, an uncertainty score acts as a proxy for the correctness likelihood of its prediction for an individual input from a specific modality. In other words, well-calibrated uncertainty indicates the expected estimation error, i.e., how far the predicted emotion is expected to lie from its ground truth. 

Given the predictions made for a set of frames from different modalities, when their uncertainty scores are well-ranked or maintain ordinality, we can effectively arrange the input unimodal frames according to their reliability for predicting a target emotion. In~\Cref{fig:trailer_fig}, we illustrate the definitions of both these constraints. \textit{It is important to note the fundamental difference between these two constraints: while the calibration constraint is applied individually for each unimodal frame, the ordinality or ranking constraint is imposed jointly for a set of frames from different modalities.}

\noindent \textbf{Calibration Constraint} -- this is imposed by regularising the unimodal context variance norms, $\| {\sigma^i_V}^2 \|_{2}$ and $\| {\sigma^i_A}^2 \|_{2}$, such that their values are strongly correlated with the correctness likelihood values of target emotion classes. In regression models, this constraint can be implemented by forcing the variance norm values to correlate with the Euclidean distance between their corresponding unimodal predictions $Y_V$ and $Y_A$ and their ground truth labels $Y^*$, as shown in~\Cref{fig:trailer_fig}. In other words, the context variance values are learnt as reliability measures indicating how far the emotion predictions are expected to lie from their ground truth labels. To impose this property on the variance values of both modalities, COLD fusion applies the following regularisation constraints, 

\begin{equation}
\begin{split}
\argmax\limits_{\sigma_V^2}  \: \: Correlation(\frac{1}{\|\sigma_V ^2\|_{2}},  \: d(Y_V, Y^*)) \\
\argmax\limits_{\sigma_A^2}  \: \: Correlation(\frac{1}{\|\sigma_A ^2\|_{2}},  \: d(Y_A, Y^*))
\end{split}
\label{eq:calib}
\end{equation}

\noindent where $d(.)$ denotes the distance function that measures the target emotion estimation error. Cross-entropy and Mean Squared Error (MSE) are used as the distance functions for the classification and regression models respectively. 

\noindent \textbf{Ordinality Constraint} -- this is applied to rank the frames of unimodal sequences, so that their uncertainty measures indicate how reliable different multimodal frames are w.r.t.\ each other. This ranking operation can be implemented as a simple ordering constraint which jointly regularises the unimodal context variance norm values, $\| {\sigma^i_V}^2 \|_{2}$ and $\| {\sigma^i_A}^2 \|_{2}$. Here, modality-wise reliability is again computed in terms of the distance values (see~\Cref{eq:calib}) between different unimodal predictions and the ground truth labels:

\begin{equation}
\begin{split}
\argmax\limits_{\sigma_V^2, \sigma_A^2} \:  \:  Correlation(Rank(\frac{1}{\|\sigma^2_V\|_{2}},  \frac{1}{\|\sigma^2_A\|_{2}}), \\  Rank(d(Y_V, Y^*), d(Y_A, Y^*)))
\end{split}
\label{eq:ordin}
\end{equation}

\subsubsection{Implementation: Calibration and Ordinality Constrained Training for Audiovisual Emotion Recognition} We train classification models of dimensional emotion recognition, in addition to the standard regression models used in the literature. In both cases, the underpinning principles of the COLD fusion are the same, but the training objective implementations differ slightly. To train the temporal context fusion models by imposing the calibration and ordinality constraints, we optimise the network to minimise a loss function composed of the following components: 

\noindent \textbf{Emotion Prediction Loss ($L_{emo}$)} is computed using the standard cross-entropy function for training the classification models. For the regression models training, similar to~\cite{Kossaifi_2020_CVPR}, we use inverse Concordance Correlation Coefficient (CCC) loss (1.0 - CCC) in addition to  MSE. This loss is computed for the predictions from unimodal ($Y_V$ and $Y_A$) and multimodal ($Y_{AV}$) branches jointly (\Cref{fig:cold_main}).

\noindent \textbf{Calibration and Ordinality Loss ($L_{CO}$)} combines the aforementioned constraints, defined in~\Cref{eq:calib} and~\Cref{eq:ordin}, into a single training objective using differentiable operations. \Cref{fig:cold_loss} shows the steps involved in implementing this component: given an input sequence with $N$ frames, we first compute their unimodal latent distributions followed by their corresponding unimodal predictions. To impose the calibration and ordinality constraints, we first compute two sets of vectors for each modality:

\textbf{Distance Vectors.} We collect the scalar distance values ($d_V^i$ and $d_A^i$) between the unimodal predictions ($Y_V^i$ and $Y_A^i$) and the ground truth labels ($Y^{i*}$) using either cross-entropy (classification) or 
MSE (regression) as the distance function. This step produces N-dimensional distance vectors, $D_V=[d_V^1, d_V^2 .., d_V^N]$ and $D_A=[d_A^1, d_A^2 .., d_A^N]$. 

\textbf{Variance-Norm Vectors.} We collect the inverted unimodal context variance norm values into another set of N-dimensional vectors, $S_V$ and $S_A$, as shown below:

\begin{equation}
    \begin{aligned}
    S_V=[\dfrac{1}{\| \sigma^1_V \|_{2}}, \dfrac{1}{\| \sigma^2_V \|_{2}}, .., \dfrac{1}{\| \sigma^N_V \|_{2}}] \\ 
    S_A=[\dfrac{1}{\| \sigma^1_A \|_{2}}, \dfrac{1}{\| \sigma^2_A \|_{2}}, .., \dfrac{1}{\| \sigma^N_A \|_{2}}].
    \end{aligned}
    \label{eq:var_vecs}
\end{equation}
 
\textbf{Softmax Distributional Matching for Calibration and Ordinal Ranking.} Note that the distance vectors and variance-norm vectors contain scalar values that summarise the properties of different embedding spaces, emotion labels, and temporal context, respectively. Hence, we assume that matching their properties by imposing the calibration and ordinality constraints directly in their original spaces, is not optimal. For this reason, as illustrated in~\Cref{fig:cold_loss}, we first apply the softmax operation on the distance vectors and variance-norm vectors separately to generate the softmax distributions. Then, we impose the calibration and ordinality constraints by minimising the mismatch between softmax distributions of the variance-norm vectors and distance vectors. This approach to calibration and ordinality loss computation based on soft-ranking is inspired by~\cite{bruch2019analysis} in which softmax cross-entropy is used for ordinal regression.

As~\Cref{fig:cold_loss} shows, in both intramodal and crossmodal settings, we compute the softmax distributions of distance vectors ({$P_{D_{V}}$, $P_{D_{A}}$, and $P_{D_{AV}}$}) and variance-norm vectors ({$P_{S_{V}}$, $P_{S_{A}}$, and $P_{S_{AV}}$}). Note that in the crossmodal case, we first concatenate the audio and visual distance vectors and variance-norm vectors separately, i.e.,  $D_{AV}=[d_A^1, d_V^1, .., d_A^N, d_V^N]$ and $S_{AV}=[s_A^1,s_V^1, ..., s_A^K,s_V^N]$. Then, we apply the softmax operation on the concatenated list which is $2N$ dimensional. Thus, the crossmodal softmax distributions capture the relative measures across both modalities. Now, to impose the calibration constraint, we minimise the KL divergence (both forward and backward) between the distance distributions and variance-norm distributions in both intramodal and crossmodal settings, as shown below:

\begin{equation}
    \begin{aligned}
    L_{CO} = KL(P_{D} || P_{S}) + KL(P_{S} || P_{D}),
    \end{aligned}
    \label{eq:L_CO}
\end{equation}

\noindent where $P_D$ represents $P_{D_{V}}$ and $P_{D_{A}}$, and $P_S$ represents $P_{S_{V}}$ and $P_{S_{A}}$ in the intramodal loss computation. In the crossmodal case, $P_D$ and $P_S$ denote $P_{D_{AV}}$ and $P_{S_{AV}}$, respectively.

\noindent \textbf{Variance Regularisation Loss ($L_{regu}$).} Prior works~\cite{chang2020data,sanchez2021affective} on latent distribution learning in high-dimensional input spaces such as images, have reported that the variance collapse is a commonly encountered problem. Variance collapse occurs mainly because the network is encouraged to predict small variance $\sigma^2$ values to suppress the unstable gradients that arise while training the latent distribution models using Stochastic Gradient Descent. To prevent this problem, we include the regularisation term proposed in~\cite{chang2020data} in the training objective:

\begin{equation}
    \begin{aligned}
    L_{regu} = KL(\mathcal{N}(\mu,\,\sigma^2) || \mathcal{N}(\epsilon,\,\textbf{I})) \\
    = -\dfrac{1}{2} (1 + log\sigma^2 - \mu^2 - \sigma^2),
    \end{aligned}
    \label{eq:L_regu}
\end{equation}

\noindent where $\epsilon$ and $I$ denote the mean vector and an identity variance matrix respectively. Note that this regularisation term is applied to the audio and visual distributions, separately.

In summary, the COLD fusion training objective composed of the above-discussed loss components, is as follows:

\begin{equation}
    \begin{aligned}
    L_{total} = L_{emo} + \lambda_{CO_V} \cdot L_{CO_V} + \lambda_{CO_{A}} \cdot  L_{CO_A} + \\
    \lambda_{CO_{AV}} \cdot L_{CO_{AV}} + \lambda_R \cdot L_{regu},
    \end{aligned}
    \label{eq:loss_main}
\end{equation}

\noindent where $\lambda_{{CO}_V}$ (for visual-only), $\lambda_{{CO}_A}$ (for audio-only), $\lambda_{{CO}_{AV}}$ (for audio and visual combined), and $\lambda_R$ (for regularisation) are the optimisation hyperparameters that control the strength of each regularisation constraint.

\section{Experiments}

We first discuss the details of dimensional \ch{and categorical} emotion datasets used for evaluating the proposed COLD fusion model. Detailed information about each dataset can be found in~\cite{ringeval2019avec,busso2008iemocap,zadeh2018multimodal}. Then, we discuss the regression and classification formulations of emotion recognition and \ch{the evaluation metrics used for dimensional and categorical emotion tasks}, along with a standard uncertainty calibration error metric that applies to the classification models. Finally, we present the details of the network architectures, fusion model implementations, and their optimisation.

\subsection{Datasets}
{\subsubsection{Dimensional Emotion Recognition}
\noindent \textbf{For spontaneous} dimensional emotion recognition, we used the AVEC 2019 CES challenge corpus~\cite{ringeval2019avec} which is designed for in-the-wild emotion recognition in cross-cultural settings as part of the SEWA project~\cite{kossaifi2019sewa}. This corpus is composed of 8.5 hours of audiovisual recordings collected from German, Hungarian, and Chinese participants. All videos in this corpus are annotated with continuous-valued valence and arousal labels in the range [-1, 1]. Note that the train and validation partitions are composed of only German and Hungarian cultures. As the labels for the test set (which has the Chinese culture in addition) are not publicly available, we report results on the validation set.  

\noindent \textbf{For acted} emotion recognition, the Interactive Emotional Dyadic Motion Capture (IEMOCAP) dataset~\cite{busso2008iemocap} is used. This dataset constitutes 12 hours of audiovisual data annotated with utterance-level labels of valence and arousal. Here, we normalised the original emotion labels to the range [-1, 1]. Among the available five sessions in this corpus, we used the first four sessions' data for training. Note that the COLD fusion model training involves tuning of multiple regularisation constraints (\Cref{eq:loss_main}). Thus, the usual 5-fold cross-validation evaluation is found to be computationally expensive as it requires the values of $\lambda_{{CO}_V}$, $\lambda_{{CO}_A}$, $\lambda_{{CO}_{AV}}$, and $\lambda_R$ to be tuned for every fold. For this reason, we used the speaker-independent partitions of the fifth session as validation and test sets, the same as the first fold's validation and test sets used in the existing works (e.g.~\cite{zhao2018exploring,tripathi2018multi}) that apply 5-fold cross-validation.   

On both the emotion datasets we trained and evaluated our audiovisual fusion models in regression as well as classification settings. To train the regression models, we directly used the continuous-valued labels as targets in the range [-1, 1].  For classification, we first mapped the continuous emotion values to three different classes for valence (positive, neutral, negative) and arousal (high, neutral, low) individually. For this binning, we chose the thresholds of -0.05 and 0.05 to draw the boundaries between the three above-mentioned bins. We adjusted the binning thresholds and picked the aforementioned values, to minimise the imbalances in the resultant class-wise label distributions.  

\noindent \textbf{Addressing Imbalanced Emotion Class Label Distributions.} Despite carefully tuning the binning thresholds, class-wise label distributions of the dimensional emotion datasets still have significant imbalances, as shown in~\Cref{fig:label_dists}. To mitigate the effect of this problem, we applied two general techniques while training the classification models: a. non-uniform sampling of the training instances for different classes and b. class-weighted cross-entropy loss. In the former, we modified the sampling criteria to over sample for the minority classes and under sample for the majority classes based on the number of examples available for each class in the train set. In the latter technique, we divided the cross-entropy loss values for different classes by their relative  bin size (in the train set).

\begin{figure}
    \centering
    \includegraphics[width=0.95\linewidth]{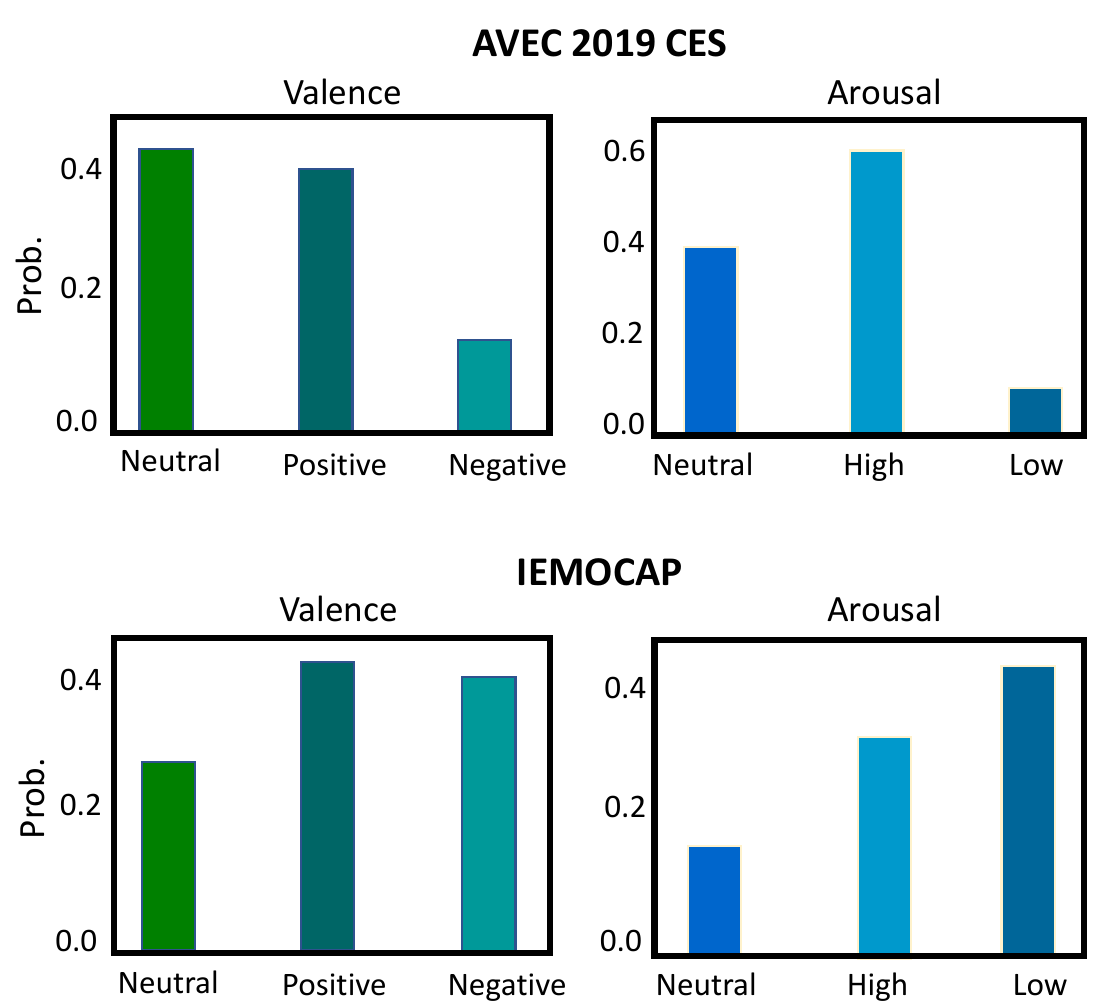}
    \caption{Class imbalances in the distributions of valence and arousal labels prepared for 3-way classification on the AVEC 2019 CES and IEMOCAP datasets.} 
    \label{fig:label_dists}
\end{figure}

\subsubsection{Categorical Emotion Recognition}
\noindent \textbf{For spontaneous} categorical emotion recognition, we used the CMU-MOSEI dataset~\cite{zadeh2018multimodal}, a large-scale dataset for in-the-wild multimodal emotion recognition. This dataset is composed of $23,453$ video utterances collected from YouTube monologues of 1000 distinct speakers. Each utterance contains three modalities: image sequences sampled at 30Hz, audio waveforms with a sample rate of 44.1kHz, and their corresponding text transcripts. All the utterances are manually annotated with 6 categorical emotions: \emph{angry}, \emph{disgust}, \emph{fear}, \emph{happy}, \emph{sad}, and \emph{surprise}. Here, we used the same training, validation and test partitions that are provided as part of the CMU-Multimodal Software Development Kit\footnote{\url{https://github.com/A2Zadeh/CMU-MultimodalSDK}}.

\noindent \textbf{For acted} categorical emotion recognition, we used the IEMOCAP dataset with the labels of six basic emotions: \emph{neutral}, \emph{angry}, \emph{happy}, \emph{sad}, \emph{excited} and \emph{frustrated}. Following the existing works ~\cite{dai2021multimodal,dai2020modality}, we used a pre-processed version of this dataset that contains $7,380$ utterances, in which each utterance contains an image sequence sampled at 30Hz, an audio waveform sampled at 16kHz, and its text transcript. We followed the same training (70\%), validation (10\%) and test (20\%) splits used in prior works (e.g.~\cite{dai2021multimodal}).

\subsection{Evaluation Metrics}
\label{sec:eval-metr}

\noindent{\textbf{Regression}} models' performance is measured using Lin’s Concordance Correlation Coefficient (CCC)~\cite{lawrence1989concordance} between the predicted emotions $y^o$ and their ground truth labels $y^*$   
\begin{equation}
    CCC = \frac{\rho_{y^*y^o} . \sigma_{y^*} . \sigma_{y^o}}{{(\mu_{y^*}-\mu_{y^o})}^2 + {\sigma_{y^*}}^2 + {\sigma_{y^o}}^2},
    \label{eq:ccc}
\end{equation}
where $\rho_{y^*y^o}$ denotes the Pearson's coefficient of correlation between $y^*$ and $y^o$, and $(\mu_{y^*}, \mu_{y^o})$ and $(\sigma_{y^*}, \sigma_{y^o})$ denote their mean and standard deviation values, respectively. 

\noindent{\textbf{Classification}} models of dimensional emotions are evaluated using precision, recall, and F1 score. \ch{Given the imbalanced emotion class distributions (see~\Cref{fig:label_dists}), for these three metrics we report unweighted or macro averaged values of the three emotion classes, so that the average values are not biased towards the most dominant classes.}  For evaluating the categorical emotion models, following prior works~\cite{dai2020modality,dai2021multimodal, li2022amoa,li2023qap}, we used (a) the accuracy and F1 score metrics for IEMOCAP and (b) the weighted accuracy and F1 score for CMU-MOSEI.

\noindent \textbf{Uncertainty Calibration Errors} of the classification models are measured to analyse the deviations between the true class likelihoods $p$ and the predicted class confidence estimates $\hat{p}$. Reliability diagrams~\cite{guo2017calibration} are used as empirical approximations to visually represent the confidence calibration errors. For plotting these diagrams, first, the accuracy and confidence axes are binned into equally-sized intervals and then, for each interval mean accuracy values are plotted against their corresponding mean confidence scores. For a perfectly calibrated model, the reliability diagram is supposed to be an identity function, i.e., accuracy and confidence should have the same values. Expected Calibration Error (ECE), a scalar summary statistic of the reliability diagram, computes the weighted average of calibration errors over all the intervals in a reliability diagram.

\begin{equation}
    ECE =  \sum_{m=1}^{M}  \frac{|I_m|}{N}  | Acc(I_m) - Conf(I_m) |,
\end{equation}

where $I_m$ denotes the $m^{th}$ interval, $M$ is the total number of intervals, and $N$ is the total number of samples.

\subsection{Network Architectures}

\subsubsection{Feature Extraction for Dimensional Emotion Models}
\noindent \textbf{Visual CNN Backbone.} EmoFAN~\cite{yang2020fan}, a 2D CNN proposed recently for facial feature extraction, is proven highly efficient by building on hour-glass-based network architectures. This CNN backbone, pretrained on 2D face alignment task, has been found very efficient for transfer learning tasks~\cite{toisoul2021estimation,ntinou2021transfer}. We used its pretrained model\footnote{Pretrained models of Toisoul et al.~\cite{toisoul2021estimation} are available at \url{https://github.com/face-analysis/emonet}} on image-based emotion recognition on the AffectNet dataset~\cite{mollahosseini2017affectnet}. Using this backbone, we extracted a 512D feature vector per frame. 

\noindent \textbf{Audio CNN Backbone.} We adopted a 2D CNN backbone proposed in~\cite{chen2019efficient} for extracting speech signal features in an end-to-end fashion. Here, we applied a VGGish~\cite{hershey2017cnn} pre-trained module to 2D Mel-spectrograms that are derived by setting the hop size and window length values to $0.1$\,s and $1$\,s respectively.  Similar to~\cite{chen2019efficient}, we fine-tuned only the last two fully connected layers of this VGGish module. To differentiate the interlocutor's information from that of the target speaker, we implemented the feature dimensionality-doubling technique proposed in~\cite{chen2017multimodal}. 

\noindent \textbf{Data Augmentation.} We applied strong data augmentation techniques to the audiovisual inputs to minimise the overfitting problem. It is important to note that under heavy overfitting, the COLD loss function (\Cref{eq:L_CO}) may collapse since the calibration and ordinality constraints rely on the prediction errors of the training instances. For face image data, we applied horizontal flipping with the probability set to 0.5, random scaling by a factor of $0.25$, random translation by +/- $30$ pixels, and random rotation by $30^{\circ}$. In the audio case, we applied SpecAugment~\cite{park2019specaugment} which directly augments the 2D spectrogram itself, instead of its original 1D waveform. Here, we applied the standard SpecAugment operations: time warping, frequency masking and time masking, with their order defined arbitrarily. The parameters\footnote{$\omega$ -- warping length, $f$ -- number of consecutive Mel frequency channels masked, $t$ -- number of consecutive time steps masked} of time warping ($\omega$), frequency masking ($f$), and time masking ($t$) are chosen from different uniform distributions in the range [0, 50], [0,27], and [0,40] respectively.

\subsubsection{\ch{Feature Extraction for Categorical Emotion Models}}
Following the existing works~\cite{dai2020modality,dai2021multimodal}, we applied the early-stage feature extraction on the aligned multimodal data. The visual features containing 35 facial action units are extracted using Facet\footnote{
iMotions. Facial expression analysis, 2017.}. The audio features extracted using COVAREP~\cite{degottex2014covarep} contain glottal source parameters, Mel-frequency cepstral coefficients, etc. Similar to the prior works~\cite{dai2020modality,dai2021multimodal}, we used 74-dimensional and 144-dimensional audio features for CMU-MOSEI and IEMOCAP datasets, respectively. The text feature vectors with 300 dimensions are prepared by tokenising the text data at word level and then extracting their GLoVE~\cite{pennington2014glove} (\footnote{glove.840B.300d: \url{https://nlp.stanford.edu/projects/glove/}})  embeddings. 

\subsubsection{Temporal Networks}
\textbf{In dimensional emotion recognition} models, the temporal networks are stacked on top of the unimodal CNN backbones to model the temporal dynamics and integrate the multimodal affect information. Note that all the fusion models evaluated in this work follow different temporal network implementations. However, all the temporal networks have the following GRU block in common: a 2-layer bidirectional GRU module  followed by a fully connected (FC) output layer. This GRU block contains 256 hidden units with the dropout value set to 0.5. The number of GRU blocks and their input-output dimensionality vary across different fusion models, as discussed below.  

In feature fusion, a single GRU+FC block is used to process the input feature sequence that is prepared via frame-wise concatenation of the unimodal embeddings, whereas, in the prediction fusion, different unimodal temporal models (GRU+FC) are applied separately, and their output softmax label distributions are aggregated into the final predictions. The context fusion implementation has two different GRU blocks, but a common FC layer. As shown in~\Cref{fig:cold_main}, COLD fusion is similar to the context fusion, but with the GRU block's output layer modified to predict the mean and variance vectors. Note that we trained the unimodal output branches simultaneously along with the fusion branch in all the multimodal models (see~\Cref{fig:cold_main}). 

\textbf{In categorical emotion recognition} models, the pre-extracted visual, audio, and text features are directly fed into their corresponding temporal networks, which are composed of the same GRU+FC blocks used in the dimensional emotion models. Except for the number of input units, which depend on the input feature dimensionality, all the network parameters are the same in both the cases. In the COLD fusion module, due to the presence of the third modality (i.e. text features) present in the categorical emotion models,~\Cref{eq:fusion_weights} is modified to accommodate three modalities and the calibration and ordinal constraints, \Cref{eq:calib} and \Cref{eq:ordin}, are modified to compute pair-wise correlations for the six possible combinations of the audio, visual and text modalities.

\subsection{Optimisation Details}  The batch size, learning rate, and weight decay values chosen for training  all these models are 4, 5e-3, and 1e-4, respectively. For tuning the learning rate, we used Cosine annealing coupled with warm restarts~\cite{loshchilov2016sgdr} (the number of epochs for the first restart set to 1 and the multiplication factor set to 2). We used Adam optimiser~\cite{kingma2014adam} for training all the models evaluated in this work.

\textbf{For dimensional emotion recognition,} we used  input sequences of 30 seconds duration with per-frame and per-sequence targets on the AVEC 2019 and IEMOCAP datasets respectively. The visual and audio backbones and all the fusion models are trained by jointly minimising the CCC loss~\cite{Kossaifi_2020_CVPR} and mean squared error for the regression task and class-weighted cross-entropy loss for the classification task. For finding the optimal values of hyper-parameters, we used the IEMOCAP validation set and the same optimal values are applied to the models trained on the AVEC 2019 corpus. The hyper-parameter values in the loss function (\Cref{eq:L_CO}) are tuned on the logarithmic scale in the range [1e-5, 1e+5] using RayTune~\cite{liaw2018tune}. Based on the IEMOCAP validation set performance, the following values are found to be optimal: 1e-3 for $\lambda_{CO_{V}}$, $\lambda_{CO_{A}}$ and $\lambda_{CO_{AV}}$, and 1e-4 for $\lambda_{R}$. We applied the same hyperparameter values to the models trained on the AVEC 2019 corpus as well. 

\textbf{For categorical emotion recognition,} we used sequences of 100 frames. The temporal networks are trained using the standard cross-entropy loss. The hyper-parameters are tuned separately on the validation sets of CMU-MOSEI and IEMOCAP. The following values are found to be optimal: 1e-2 and 5e-3 for \{$\lambda_{CO_{V}}$, $\lambda_{CO_{A}}$, and $\lambda_{CO_{AV}}$\} on CMU-MOSEI and IEMOCAP respectively and, 1e-4 and 5e-5 for $\lambda_{R}$ on CMU-MOSEI and IEMOCAP respectively.

\section{Results and Discussion}

We first present the results of dimensional \ch{and categorical} emotion recognition models based on different audiovisual fusion techniques. By inducing visual noise through face masking, we investigate the robustness of the proposed COLD fusion compared to the standard fusion baselines. Then, we analyse the uncertainty calibration performance of the COLD fusion model, particularly in classification settings. Finally, a qualitative analysis of modality-wise fusion weights is presented to demonstrate the calibration and ordinal ranking proprieties of the COLD fusion model. 

\begin{table}
    \centering
    \begingroup
    \renewcommand{\arraystretch}{1.3}        
    \begin{tabular}{l r r r}
    \toprule
    & \textbf{Valence} & \textbf{Arousal} & \textbf{Avg.} \\
    \textbf{Model} & \textbf{CCC $\uparrow$} & \textbf{CCC $\uparrow$} & \textbf{CCC $\uparrow$} \\
    \midrule
    AVEC CES Winners:Aud~\cite{zhao2019adversarial} & 0.388 & 0.518 & 0.453 \\
    Aud-branch & 0.369 & 0.465 & 0.417 \\
    AVEC CES Winners:Vis~\cite{zhao2019adversarial} & 0.579 & 0.594 & 0.586 \\
    Vis-branch  & 0.511 & 0.514 & 0.512 \\
    \midrule
    AVEC CES Winners:AV Fusion~\cite{zhao2019adversarial} & \textbf{0.614} & 0.645 & 0.629 \\
    AV Feature Fusion & 0.515 & 0.509 & 0.512 \\
    AV Prediction Fusion  & 0.552 & 0.617 & 0.584 \\
    \textit{AV Context Fusion}  & 0.578 & 0.620 & 0.599 \\
    \textit{AV COLD Fusion} & 0.611 & \textbf{0.661} & \textbf{0.636} \\    
    \bottomrule
    \end{tabular}
    \caption{Dimensional emotion \textit{regression} results on the \textbf{AVEC 2019 CES validation set} (CCC: Concordance Correlation Coefficient).}
    \label{tab:avec19_reg}
    \endgroup
\end{table}

\begin{table}
    \centering
    \begingroup
    \renewcommand{\arraystretch}{1.3}        
    \begin{tabular}{l r r r}
    \toprule
    & \textbf{Valence} & \textbf{Arousal} & \textbf{Avg.} \\
    \textbf{Model} & \textbf{CCC $\uparrow$} & \textbf{CCC $\uparrow$} & \textbf{CCC $\uparrow$} \\
    \midrule
    Aud-branch & 0.694 & 0.453 & 0.573 \\
    Vis-branch  & 0.496 & 0.355 & 0.425\\
    \midrule
    AV Feature Fusion & 0.568 & 0.390 & 0.479 \\
    AV Prediction Fusion  & 0.696 & 0.481 & 0.578 \\
    \textit{AV Context Fusion}  & 0.690 & 0.487 & 0.588 \\
    \textit{AV COLD Fusion} & \textbf{0.723} & \textbf{0.504} & \textbf{0.613} \\    
    \bottomrule
    \end{tabular}
    \caption{Dimensional emotion \textit{regression} results  on the \textbf{IEMOCAP test set} (CCC: Concordance Correlation Coefficient).}
    \label{tab:iemocap_reg}
    \endgroup
\end{table}

\begin{figure*}
    \centering
    \includegraphics[width=1.01\linewidth]{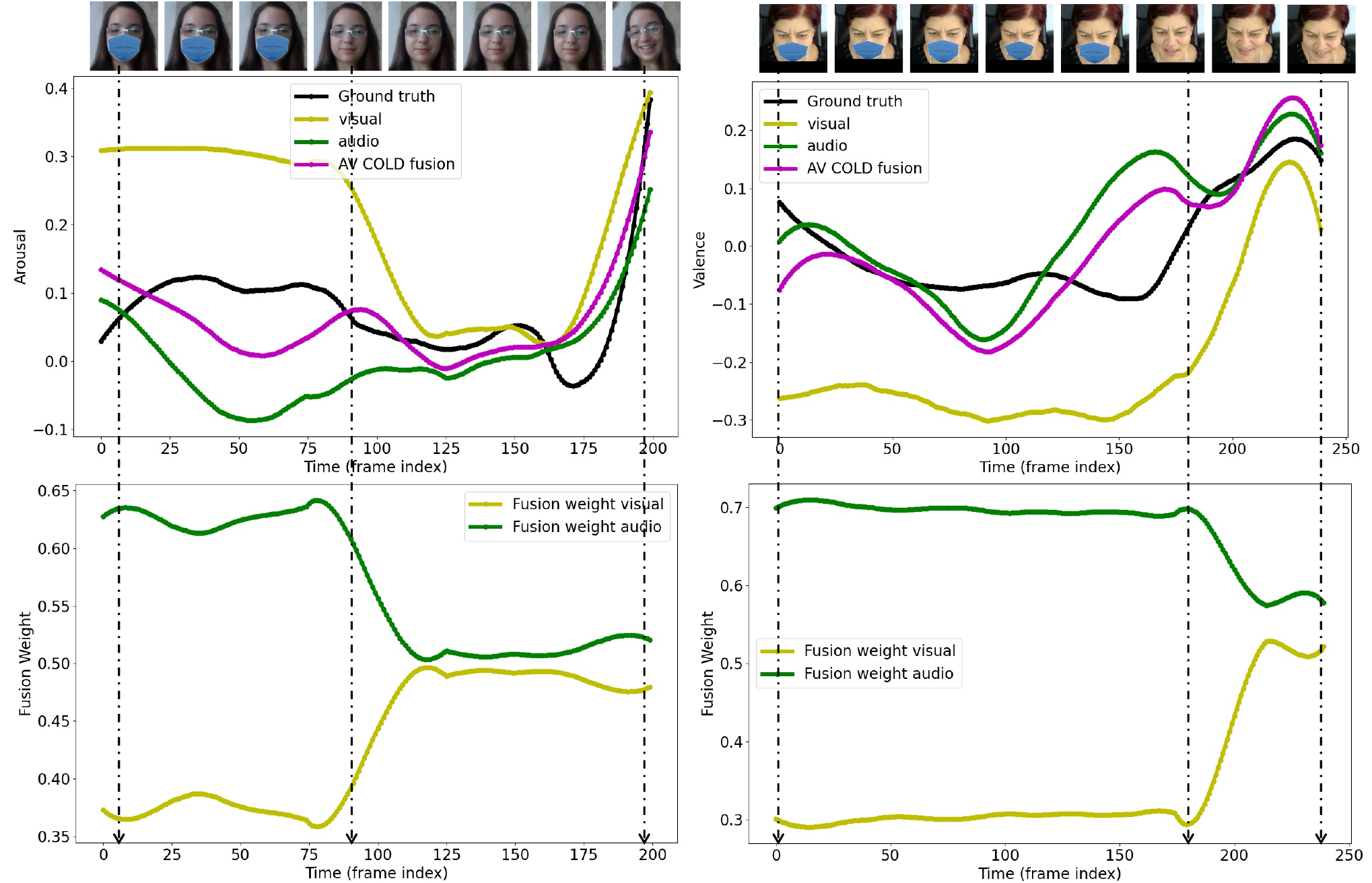}
    \caption{Dynamic adaptation of COLD fusion weights when presented with novel noise patterns induced into the visual inputs: At test time, face masking is applied to randomly chosen consecutive frames in the AVEC 2019 CES validation examples. When the visual modality is noisy, i.e., containing faces with masks, AV COLD fusion output relies more on the audio modality (note the gaps between visual predictions and AV COLD fusion predictions, and modality-wise fusion weights). After removing the face masks, the fusion weight values adapt accordingly, hence, the fusion outputs.}
    \label{fig:masking_res}
\end{figure*}

\begin{figure}
    \centering
    \includegraphics[width=1\linewidth]{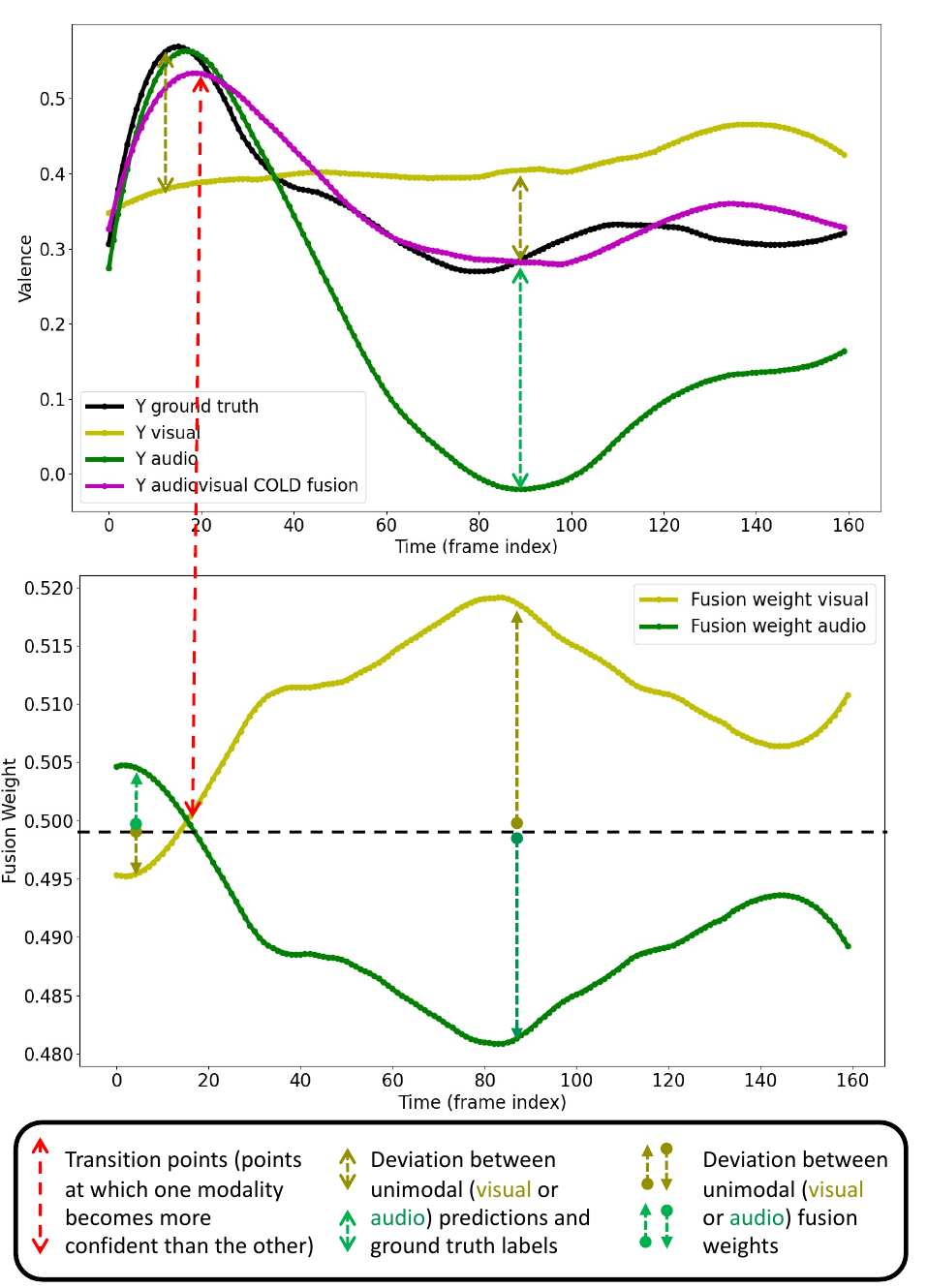}
    \caption{Emotion predictions on an example from the AVEC 2019 CES validation set: Unimodal and multimodal valence predictions, and their uncertainty-based fusion weights estimated by the AV COLD fusion predictions. Note that fusion weights of the audio and visual modalities demonstrate (a) the calibration property -- how far their corresponding unimodal predictions are from the ground truth ratings and (b) the ordinal ranking property -- how well they can order the audio and visual modalities in terms of their reliability.} 
    \label{fig:time_plots}
\end{figure}

\subsection{Dimensional Emotion Recognition Results}  
 
\noindent \textbf{Regression} performance of different unimodal (Aud-branch and Vis-branch) and multimodal (AV) predictions are presented in~\Cref{tab:avec19_reg} and~\Cref{tab:iemocap_reg} for the AVEC 2019 CES (spontaneous emotion recognition) and IEMOCAP (acted emotion recognition) corpora, respectively. In both cases, COLD fusion consistently outperformed the standard fusion baselines (feature, prediction and context) as well as the unimodal results. When compared to the best performing CNN+RNN fusion baselines COLD fusion achieved $\sim$6\% average relative improvement.

Compared to the winners of the AVEC 2019 challenge, Zhao et al.~\cite{zhao2019adversarial}, COLD fusion performs well in terms of arousal and mean CCC scores. However, it is slightly worse in the case of valence CCC. Note that Zhao et al.~\cite{zhao2019adversarial} use a domain adaptation technique to cope with the cross-cultural variations in audiovisual emotion expressions. However, our focus is not on coping with the cross-cultural variations, but primarily on improving the fusion performance. It is important to note that our fusion technique is, in principle, complementary to the domain adaptation used in~\cite{zhao2019adversarial}. More advanced temporal models such as Affective Processes~\cite{tellamekala2022modelling,sanchez2021affective,mani2021stochastic} demonstrated superior generalisation performance than the RNNs in recent years. However, since this work mainly focuses on capturing temporal uncertainty for model-agnostic fusion based on simple CNN+RNN formulations, such complex temporal models based on APs are not included in this comparison, to not clutter the analysis of standard model-agnostic fusion methods presented here.

\ch{In Appendix.~\ref{app:transformer}, we compare the proposed COLD fusion and a multimodal Transformer baseline~\cite{tsai2019multimodal} on the AVEC 2019 dimensional emotion regression task. Here also, COLD fusion clearly outperformed the transformer baseline by a noticeable margin, especially in arousal prediction.}

\ch{Appendix.~\ref{app:ablation} presents an ablation study of different components in the COLD fusion formulation, by nullifying different hyperparameters to modify the COLD training objective (\Cref{eq:loss_main}). These results, as shown in \Cref{tab:ablation}, show the importance of calibration, ordinal, and variance regularisation constraints to the overall performance improvements achieved by COLD fusion. In Appendix.~\ref{app:stat_sig}, we present the results of statistical significance tests, further validating the improvements achieved by COLD fusion over the standard fusion baselines.}  \\

\begin{table}[]
    \centering
    \begingroup
    \renewcommand{\arraystretch}{1.3}    
    \begin{tabular}{l  r r r  r r r }
    \toprule
        & \multicolumn{3}{c}{\textbf{Valence}} & \multicolumn{3}{c}{\textbf{Arousal}} \\
        \cmidrule(lr){2-4} \cmidrule(lr){5-7}
          \textbf{Model} &  \textbf{P$\uparrow$} & \textbf{R$\uparrow$} & \textbf{F1$\uparrow$} &  \textbf{P$\uparrow$} & \textbf{R$\uparrow$} & \textbf{F1$\uparrow$} \\
        \hline
    Aud-branch & 68.3 & 48.2 & 56.6 &                   74.3 & 50.2 & 59.9 \\
    Vis-branch  & 70.9 & 58.1 & 63.9 &                 76.8 & 70.3 & 73.4 \\
    AV Feature Fusion & 67.8 &  60.2 & 63.8 &            73.4 & 68.2 & 70.7 \\
    AV Prediction Fusion  & 68.9 & 60.5 & 64.4 &  77.0 & 69.4 & 73.0 \\
    \textit{AV Context Fusion}  & 75.0 & 60.6 & 67.0 &            77.1 & 71.1 & 73.9 \\
    \textit{AV COLD Fusion} & \textbf{76.8} & \textbf{62.4} & \textbf{68.9} &      \textbf{79.5} & \textbf{74.0} & \textbf{76.5} \\
    \bottomrule
    \end{tabular}
    \caption{Dimensional emotion \textit{3-way classification} results (P: Precision, R: Recall, F1: F1 score) on the \textbf{AVEC 2019 CES validation set}.}
    \label{tab:avec19_classif}
    \endgroup
\end{table}

\begin{table}[]
    \centering
    \begingroup
    \renewcommand{\arraystretch}{1.3}    
    \begin{tabular}{l  r r r  r r r }
    \toprule
        & \multicolumn{3}{c}{\textbf{Valence}} & \multicolumn{3}{c}{\textbf{Arousal}} \\
    \cmidrule(lr){2-4} \cmidrule(lr){5-7}
    \textbf{Model} &  \textbf{P$\uparrow$} & \textbf{R$\uparrow$} & \textbf{F1$\uparrow$} &  \textbf{P$\uparrow$} & \textbf{R$\uparrow$} & \textbf{F1$\uparrow$} \\
        \hline
    Aud-branch &                 55.5 & 55.3 & 55.4 & 62.4 & 55.1 & 58.5 \\
    Vis-branch &                 53.7 & 48.7 & 51.0 & 56.8 & 49.2 & 52.7 \\
    AV Feature Fusion &          62.8 & 47.8 & 54.3 & 58.4 & 54.2 & 56.2 \\
    AV Prediction Fusion  &           64.5 & 57.3 & 60.7 & 58.9 & 57.0 & 57.9 \\
    \textit{AV Context Fusion} & 64.3 & \textbf{61.6} & 62.9 & 61.8 & 55.6 & 58.6\\
    \textit{AV COLD Fusion} &    \textbf{66.3} & 61.5 & \textbf{63.8} & \textbf{64.9} & \textbf{59.5} & \textbf{62.1} \\
    \bottomrule
    \end{tabular}
    \caption{Dimensional emotion \textit{3-way classification} results (P: Precision, R: Recall, F1: F1 score) on the \textbf{IEMOCAP test set}.}
    \label{tab:iemocap_classif}
    \endgroup
\end{table}

\noindent \textbf{Classification } performance on the AVEC 2019 CES and IEMOCAP corpora is presented in~\Cref{tab:avec19_classif,tab:iemocap_classif}. Similar to the regression results, COLD fusion demonstrates superior emotion classification results on both datasets. Note that here, we pose the original regression problem as a 3-way classification problem by discretising the continuous emotion labels. For this reason, we do not have any existing benchmarks for comparison in this particular classification setting. Nevertheless, the performance improvements achieved by the COLD fusion are consistent  for both valence and arousal in terms of all three metrics, except for the valence recall on IEMOCAP. \\

\noindent \textbf{Unimodal Performance Analysis.} It is interesting to note that in the AVEC 2019 case, the visual modality (Vis-branch) has a considerably better performance compared to the audio modality (Aud-branch), while it is vice versa in the case of the IEMOCAP dataset. This discrepancy may be due to the difference in the quality of the video data in terms of face image resolution. Despite such dataset-specific differences, our COLD fusion technique shows consistent performance improvements in the multimodal classification and regression settings for both datasets. \\

\noindent \textbf{Analysis of Fusion Baselines.} 
Among the fusion methods that we evaluated here, temporal context or simply context fusion is found to be the second-best performing method after the proposed COLD fusion, on both datasets. Note that here, the temporal context refers to the output of the unimodal GRU block, and unimodal predictions are generated by applying a shallow fully connected network to the unimodal context vector. Thus, the context vectors can be viewed as higher-dimensional descriptors of the final unimodal predictions. Based on this assumption, in theory, the performance of context fusion is bound to be either better or at least as good as the prediction fusion, justifying the trends observed in our experimental results.

We notice that the feature fusion performance is inferior to all the remaining fusion techniques, and prediction fusion performs better than feature fusion. This result is consistent with an observation that prediction fusion achieves better results compared to feature fusion in general, as reported in the existing multimodal affect recognition literature~\cite{ringeval2015prediction}. It is worth noting that the results of feature fusion are worse than that of the best performing unimodal models on both datasets, i.e., the visual (Vis-branch) model on AVEC 2019 and the audio (Aud-branch) model on IEMOCAP. This performance degradation may be due to not explicitly correcting the temporal misalignment effects~\cite{lingenfelser2016asynchronous}, which are heuristically derived in general~\cite{ringeval2019avec}. This result indicates that integrating multimodal emotion information at the feature-level or frame-level could be suboptimal most likely due to the temporal misalignment issues, given that continuous emotion information is expressed in the audiovisual modalities at different frame rates~\cite{rouast2019deep,tzirakis2017end}. \\

\begin{table}[]
    \centering
    \begingroup
    \renewcommand{\arraystretch}{1.3}    
    \begin{tabular}{l  r r  r r }
    \toprule
        & \multicolumn{2}{c}{\textbf{Valence ECE $\downarrow$}} & \multicolumn{2}{c}{\textbf{ Arousal ECE $\downarrow$}} \\
          \textbf{Model} &  \textbf{BTS} & \textbf{ATS} &  \textbf{BTS} & \textbf{ATS}  \\
        \hline
    Aud-branch & 13.6e-2 & 6.3e-2 & 3.2e-2 & 2.8e-2 \\
    Vis-branch & 8.9e-2 & 7.1e-2 & 12.6e-2 & 3.1e-2 \\
    AV Feature Fusion & 6.1e-2 & 5.0e-2 & 5.5e-2 & 3.3e-2 \\
    AV Prediction Fusion  & 8.7e-2 & 5.1e-2 & 2.5e-2 & 2.6e-2 \\
    \textit{AV Context Fusion} & 6.9e-2 & \textbf{4.0e-2} & 6.3e-2 & 3.0e-2 \\
    \textit{AV COLD Fusion} & \textbf{3.7e-2} & 4.3e-2 & \textbf{1.3e-2} & \textbf{0.9e-2} \\
    \bottomrule
    \end{tabular}
    \caption{Dimensional emotion classification \textit{calibration} results on the \textbf{AVEC 2019 CES validation set}  (ECE: Expected Calibration Error, BTS: Before Temperature Scaling, ATS: After Temperature Scaling).}
    \label{tab:avec19_calib}
    \endgroup
\end{table}

\begin{table}[]
    \centering
    \begingroup
    \renewcommand{\arraystretch}{1.3}    
    \begin{tabular}{l  r r  r r }
    \toprule
        & \multicolumn{2}{c}{\textbf{Valence ECE $\downarrow$}} & \multicolumn{2}{c}{\textbf{Arousal ECE $\downarrow$}} \\
          \textbf{Model} &  \textbf{BTS} & \textbf{ATS} &  \textbf{BTS} & \textbf{ATS}  \\
        \hline
    Aud-branch & 7.4e-2 & 6.0e-2 & 13.2e-2 & 8.1e-2 \\
    Vis-branch  & 8.0e-2 & 4.1e-2 & 7.4e-2 & 5.0e-2 \\
    AV Feature Fusion & 7.5e-2 & 5.6e-2 & 7.0e-2 & 5.8e-2 \\
    AV Prediction Fusion  & 8.4e-2 & 6.4e-2 & 8.4e-2 & 4.2e-2  \\
    \textit{AV Context Fusion} & 9.7e-2 & 6.6e-2 & 7.9e-2 & 6.0e-2 \\
    \textit{AV COLD Fusion} & \textbf{6.1e-2} & \textbf{3.8e-2} & \textbf{1.1e-2} & \textbf{2.0e-2} \\
    \bottomrule
    \end{tabular}
    \caption{Dimensional emotion classification \textit{calibration} results on the  \textbf{IEMOCAP test set} (ECE: Expected Calibration Error, BTS: Before Temperature Scaling, ATS: After Temperature Scaling)).}
    \label{tab:iemocap_calib}
    \endgroup
\end{table}

\noindent \textbf{Dynamic Adaptation of Fusion Weights in the Presence of Noise.} In this experiment, we aim to understand how different fusion models perform when presented with novel noise patterns at test time. By inducing noise into the visual modality through face masking, here, we investigate the performance of different fusion baselines in comparison with the COLD fusion. For this evaluation, we overlaid the face masks as external occlusions on the image sequences using the method proposed in {MaskTheFace}~\cite{anwar2020masked}\footnote{\url{https://github.com/aqeelanwar/MaskTheFace}}. We applied MaskTheFace to 50\% of the randomly chosen consecutive frames of the AVEC 2019 CES validation set sequences, as shown in~\Cref{fig:masking_res}. Note that all the fusion models evaluated here have not seen faces with masks during their training. As~\Cref{tab:face_masking_avec19} shows, in this noise-induced evaluation setup, performance drop compared to the noise-free evaluation (\Cref{tab:avec19_reg}) is considerably higher for all three fusion baselines (feature, prediction, and context) than for the COLD fusion. Furthermore, the relative performance difference between the COLD fusion and the best-performing fusion baselines is increased from $\sim$6\% in noise-free settings to $\sim$17\% in this noise-induced case.

\Cref{fig:masking_res} compares the COLD fusion predictions with the predictions from visual and audio branches, along with the inferred modality-wise fusion weight scores. We can clearly see that the visual fusion weights are much lower for the frames with masks compared to the frames without masks, and as a result, the final predictions rely more on the audio modality in the presence of visual noise. This result demonstrates the ability of COLD fusion to dynamically adjust the importance of a specific modality according to its informativeness towards recognising the target emotions. \\

\begin{table}
    \centering
    \begingroup
    \renewcommand{\arraystretch}{1.3}        
    \begin{tabular}{l r r r}
    \toprule
    & \textbf{Valence} & \textbf{Arousal} & \textbf{Avg.} \\
    \textbf{Model} & \textbf{CCC $\uparrow$} & \textbf{CCC $\uparrow$} & \textbf{CCC $\uparrow$} \\
    \midrule
    AV Feature Fusion & 0.378 & 0.351 & 0.364 \\
    AV Prediction Fusion  & 0.363 & 0.545 & 0.454 \\
    \textit{AV Context Fusion}  & 0.385 & 0.508 & 0.445  \\
    \textit{AV COLD Fusion} & \textbf{0.491} & \textbf{0.574} & \textbf{0.528}\\    
    \bottomrule
    \end{tabular}
    \caption{Impact of visual noise (external occlusions) on the AV fusion models: Dimensional emotion \textit{regression} results \emph{with 50\% of randomly chosen face images masked during evaluation (see~\Cref{fig:masking_res})} on the AVEC 2019 CES validation Set.}
    \label{tab:face_masking_avec19}
    \endgroup
\end{table}

\subsection{Categorical Emotion Recognition Results}
\label{sec:categorical_emo}

\begin{table*}[]
    \centering
    \begingroup
    \renewcommand{\arraystretch}{1.3}    
    \begin{tabular}{l  c c c c c c c c c c c c c c}
    \toprule
        & \multicolumn{2}{c}{\textbf{Angry}} & 
         \multicolumn{2}{c}{\textbf{Disgusted}} & 
         \multicolumn{2}{c}{\textbf{Fear}} & 
         \multicolumn{2}{c}{\textbf{Happy}} & 
         \multicolumn{2}{c}{\textbf{Sad}} & 
         \multicolumn{2}{c}{\textbf{Surprised}} & 
         \multicolumn{2}{c}{\textbf{Average}}  \\
    \cmidrule(lr){2-3} \cmidrule(lr){4-5} \cmidrule(lr){6-7} \cmidrule(lr){8-9} \cmidrule(lr){10-11} \cmidrule(lr){12-13} \cmidrule(lr){14-15}
    \textbf{Model} &  
    \textbf{WAc.$\uparrow$} & \textbf{F1$\uparrow$} &  
    \textbf{WAc.$\uparrow$} & \textbf{F1$\uparrow$} &  
    \textbf{WAc.$\uparrow$} & \textbf{F1$\uparrow$} &  
    \textbf{WAc.$\uparrow$} & \textbf{F1$\uparrow$} &  
    \textbf{WAc.$\uparrow$} & \textbf{F1$\uparrow$} &  
    \textbf{WAc.$\uparrow$} & \textbf{F1$\uparrow$} &  
    \textbf{WAc.$\uparrow$} & \textbf{F1$\uparrow$}    \\
    \midrule

LSTM Pred. Fusion$^\dagger$   & 64.5 & 47.1 & 70.5 & 49.8 & 61.7 & 22.2 & 61.3 & 73.2 & 63.4 & 47.2 & 57.1 & 20.6 & 63.1 & 43.3  \\
Transf. Pred. Fusion$^\dagger$ & 65.3 & 47.7 & 74.4 & 51.9 & 62.1 & 24.0 & 60.6 & 72.9 & 60.1 & 45.5 & 62.1 & 24.2 & 64.1 & 44.4 \\
EmoEmbs~\cite{dai2020modality}                   & 66.8 & \textbf{49.4} & 69.6 & 48.7 & 63.8 & 23.4 & 61.2 & 71.9 & 60.5 & 47.5 & 63.3 & 24.0 & 64.2 & 44.2 \\
MulT~\cite{tsai2019multimodal}$^\dagger$   & 64.9 & 47.5 & 71.6 & 49.3 & 62.9 & 25.3 & \textbf{67.2} & \textbf{75.4} & \textbf{64.0} & \textbf{48.3} & 61.4 & 25.6 & 65.4 & 45.2 \\
AMOA~\cite{li2022amoa}$^\dagger$                          & 66.4 & 47.5 & 74.9 & 52.2 & 62.0 & 25.1 & 62.6 & 73.4 & 63.8 & 47.2 & 64.3 & 26.5 & 65.7 & 45.3 \\
AVL Context Fusion  & 61.8 & 47.3 & \textbf{79.6} & 50.6 & 64.9 & 22.7 & 66.3 & 73.3 & 55.8 & 48.1 & 63.4 & 24.5 & 65.3 & 44.5 \\
AVL COLD Fusion      & \textbf{69.7} & 49.3 & 77.6 & \textbf{52.9} & \textbf{79.8} & \textbf{25.6} & 65.8 & 74.2 & 54.8 & 46.5 & \textbf{78.8} & \textbf{28.5} & \textbf{71.1} & \textbf{46.1} \\ 

    \bottomrule
    \end{tabular}
    \caption{\ch{Categorical emotion classification results on the \textbf{CMU-MOSEI} test set (WAc. and F1 indicate weighted accuracy and F1 scores respectively and $^\dagger$ indicates the baseline models' results from Li et al~\cite{li2023qap}).}}
    \label{tab:mosei_cat_emo_classif}
    \endgroup
\end{table*}

\begin{table*}[]
    \centering
    \begingroup
    \renewcommand{\arraystretch}{1.3}    
    \begin{tabular}{l  c c c c c c c c c c c c c c}
    \toprule
        & \multicolumn{2}{c}{\textbf{Angry}} & 
         \multicolumn{2}{c}{\textbf{Excited}} & 
         \multicolumn{2}{c}{\textbf{Frustrated}} & 
         \multicolumn{2}{c}{\textbf{Happy}} & 
         \multicolumn{2}{c}{\textbf{Neutral}} & 
         \multicolumn{2}{c}{\textbf{Sad}} & 
         \multicolumn{2}{c}{\textbf{Average}}  \\
    \cmidrule(lr){2-3} \cmidrule(lr){4-5} \cmidrule(lr){6-7} \cmidrule(lr){8-9} \cmidrule(lr){10-11} \cmidrule(lr){12-13} \cmidrule(lr){14-15}
    \textbf{Model} &  
    \textbf{Acc.$\uparrow$} & \textbf{F1$\uparrow$} &  
    \textbf{Acc.$\uparrow$} & \textbf{F1$\uparrow$} &  
    \textbf{Acc.$\uparrow$} & \textbf{F1$\uparrow$} &  
    \textbf{Acc.$\uparrow$} & \textbf{F1$\uparrow$} &  
    \textbf{Acc.$\uparrow$} & \textbf{F1$\uparrow$} &  
    \textbf{Acc.$\uparrow$} & \textbf{F1$\uparrow$} &  
    \textbf{Acc.$\uparrow$} & \textbf{F1$\uparrow$}    \\
    \midrule

    LSTM Pred. Fusion$^\dagger$   & 71.2 & 49.4 & 79.3 & 57.2 & 68.2 & 51.5 & 67.2 & 37.6 & 66.5 & 47.0 & 78.2 & 54.0 & 71.8 & 49.5 \\
    Transf. Pred. Fusion$^\dagger$ & 81.9 & 50.7 & 85.3 & 57.3 & 60.5 & 49.3 & 85.2 & 37.6 & 72.4 & 49.7 & 87.4 & 57.4 & 78.8 & 50.3 \\
    EmoEmbs~\cite{dai2020modality}                 & 65.9 & 48.9 & 73.5 & 58.3 & 68.5 & 52.0 & 69.6 & 38.3 & 73.6 & 48.7 & 80.8 & 53.0 & 72.0 & 49.8 \\
    MulT~\cite{tsai2019multimodal}$^\dagger$  & 77.9 & \textbf{60.7} & 76.9 & 58.0 & 72.4 & 57.0 & 80.0 & 46.8 & 74.9 & \textbf{53.7} & 83.5 & \textbf{65.4} & 77.6 & \textbf{56.9} \\
    AMOA~\cite{li2022amoa}$^\dagger$   & 82.5 & 53.4 & 85.8 & 57.9 & \textbf{74.4} & 56.5 & \textbf{88.6} & \textbf{47.0} & 73.2 & 49.6 & 87.8 & 64.5 & 82.1 & 54.8 \\
    AVL Context Fusion& 85.4 & 51.6 & 82.4 & 52.2 & 63.4 & 50.0 & 85.6 & 35.5 & 71.6 & 49.3 & 84.6 & 56.8 & 78.8 & 49.2 \\
    AVL COLD Fusion   & \textbf{86.5} & 58.4 & \textbf{88.3} & \textbf{61.6} & 68.6 & \textbf{57.9} & 88.1 & 43.7 & \textbf{76.6} & 48.9 & \textbf{87.9} & 60.2 & \textbf{82.7} & 55.1 \\ 

    \bottomrule
    \end{tabular}
    \caption{\ch{Categorical emotion classification results on the \textbf{IEMOCAP} test set (Acc. and F1 indicate standard accuracy and F1 scores respectively, and $^\dagger$ indicates the baseline models'  results from Li et al~\cite{li2023qap}).}}
    \label{tab:iemocap_cat_emo_classif}
    \endgroup
\end{table*}

The comparative results for the categorical emotion recognition tasks are presented in \Cref{tab:mosei_cat_emo_classif} (CMU-MOSEI) and \Cref{tab:iemocap_cat_emo_classif} (IEMOCAP). This comparison considers the following baselines: late fusion models based on LSTMs and Transformers, existing multimodal benchmarks and a SOTA model (AMOA~\cite{li2022amoa}) among the two-phase models. Note that most existing models evaluated on CMU-MOSEI and IEMOCAP take a two-phase approach to multimodal emotion recognition, in which unimodal hand-crafted feature extraction and multimodal temporal fusion are performed separately. In line with those works, we evaluated the COLD fusion using  the same two-phase architecture.

As shown in \Cref{tab:mosei_cat_emo_classif} and \Cref{tab:iemocap_cat_emo_classif}, COLD fusion achieves new SOTA performance among the two-phase models. Note that on the both datasets there is a noticeable performance difference between the context fusion and COLD fusion models, which demonstrates the importance of the proposed calibration and ordinal constraints on temporal latent distribution learning. On CMU-MOSEI, compared to the existing SOTA (AMOA), COLD fusion achieves 8.2\% and 1.8\% relative improvements in terms of the average weighted accuracy and F1 scores respectively.  On IEMOCAP, COLD fusion demonstrates the best accuracy and the second best F1 score. Here, the model with the highest F1 score is based on a multimodal transformer (Mult~\cite{tsai2019multimodal}), whereas the COLD fusion model implemented in this work uses GRUs for modelling the temporal dynamics. For further performance improvements, the proposed COLD fusion model can be integrated with transformer-based temporal models for combining the best of both worlds.

Compared to the two-phase models considered here for evaluation, some recently proposed fully end-to-end models such as the ones in \cite{dai2021multimodal, wei2022fv2es, li2023qap} demonstrated improved emotion recognition performance but at the cost of significantly increased model training complexity. Although the COLD fusion framework is not evaluated in such models in this work, its ability to achieve robust multimodal fusion can be extended to fully end-to-end models as well for additional performance gains.

To demonstrate the applicability of COLD fusion to other multimodal tasks, besides emotion recognition, we evaluated it on utterance-level multimodal (AVL) sentiment analysis tasks on the CMU-MOSEI dataset. Refer to Appendix.~\ref{app:mosei} for the sentiment classification and regression results of the COLD fusion model in comparison with the existing baselines. In this case, COLD fusion achieves competitive results compared to the best-performing baseline (MISA~\cite{hazarika2020misa}) and it shows the best results when coupled with MISA.

Overall, the multimodal categorical emotion and sentiment recognition results demonstrate the importance of learning well-calibrated and well-ranked uncertainty scores for improved multimodal fusion performance. These experiments also show that the COLD fusion formulation can be easily extended to models with more than two modalities.

\subsection{Uncertainty Calibration Performance Analysis} 
To measure the quality of uncertainty estimates, we computed Expected Calibration Error (ECE) (see~\Cref{sec:eval-metr}) values for the unimodal and multimodal emotion classification models. Note that this calibration error metric applies only to the classification settings. By computing the ECE values before and after applying temperature scaling to the softmax distributions over the predictions of each model separately, we analyse the impact of explicit uncertainty calibration (temperature scaling). We searched for an optimal temperature value in the range of $1e-2$ to $1000$ by doing a random search for $100$ iterations. Similar to the technique followed in~\cite{mukhoti2020calibrating}, we selected a temperature value that achieves the lowest ECE value on the validation set.   

It is important to consider that the COLD fusion models are trained to be implicitly calibrated (see~\Cref{eq:L_CO}) in terms of their context variance values. Thus, even before applying explicit calibration, i.e., temperature scaling, we expect the predictive uncertainty values or class-wise confidence scores of the COLD fusion models to have lower ECE values compared to the other fusion baselines.

\Cref{tab:avec19_calib} reports the ECE values for valence and arousal attributes on the AVEC 2019 corpus. For both attributes, before the application of temperature scaling, COLD fusion has the lowest calibration error when compared to the other models. After applying temperature scaling, it is obvious that the ECE values for all the models go down, and the COLD fusion still achieves the lowest error. Only in the case of valence, AV context fusion has a marginally lower ECE value compared to the COLD fusion. This minor discrepancy could be due to the random search of optimal temperature values and note that here, different models have different optimal temperature values that are tuned for valence and arousal, separately. Nevertheless, in all the remaining cases (both before and after temperature scaling), COLD fusion consistently shows lower uncertainty calibration errors w.r.t. the other fusion models. Results on the IEMOCAP corpus (see \Cref{tab:iemocap_calib}) show similar trends, validating the effectiveness of the COLD fusion approach in producing well-calibrated uncertainty estimates. \ch{To visually illustrate the uncertainty calibration performance of the COLD fusion model, Appendix.~\ref{app:reliability} compares the reliability diagrams of different unimodal and multimodal dimensional emotion classification models.}   \\

\noindent \textbf{Analysis of Audiovisual Fusion Weights. } \Cref{fig:time_plots} illustrates modality-wise fusion weights estimated by the COLD fusion model on a validation sequence taken from the AVEC 2019 corpus. Note that these fusion weights are functions of the unimodal temporal context distributions (see~\Cref{eq:fusion_weights}). In this illustration, we analyse the temporal patterns of fusion weights along with their corresponding unimodal and multimodal emotion predictions and their ground truth labels. This analysis clearly shows the well-calibrated nature of modality-wise fusion weights: when the predictions of one modality move closer to the ground truth compared to those of the other modality, the audiovisual weight values in the COLD fusion are found to be varying accordingly. From the transition points marked in~\Cref{fig:time_plots}, we can see that the fusion weights are gradually inverted, as the predictions of one modality move closer to the ground truth while the other modality predictions move further. This result validates our main hypothesis of making unimodal latent distributions calibrated and ordinal for improved fusion performance.

\section{Conclusion}
\label{sec:conc}

We proposed an uncertainty-aware multimodal fusion approach to dimensional and categorical emotion recognition from multimodal data. To capture modality-wise uncertainty w.r.t.\  predicting valence and arousal dimensions, we probabilistically modelled the unimodal temporal context by learning modality-wise latent distributions. For effective uncertainty-weighted multimodal fusion, we suggested conditioning the unimodal latent distributions such that their variance norms are learnt to be \emph{well-calibrated} and \emph{well-ranked (ordinal)}. To jointly impose these two constraints on the latent distributions, we introduced a novel softmax distributional matching loss function that encourages the uncertainty scores to be well-calibrated and well-ranked. Our novel loss function for multimodal learning is applicable to both classification and regression settings.

For example, in dimensional emotion regression tasks, COLD fusion shows $\sim$6\% average relative improvement over the best performing fusion baseline. Similarly, in the case of categorical emotion classification, COLD fusion achieves $\sim$8.2\% relative accuracy improvement over the existing state-of-the-art model. Furthermore, we assess the robustness of different fusion models at test time by inducing noise into the visual modality through face masking. With the faces masked in 50\% of the evaluation sequences, COLD fusion achieves $\sim$17\% average relative improvement over the best fusion baseline.

On spontaneous and acted emotion recognition tasks \ch{(in both dimensional and categorical emotion cases)}, our proposed uncertainty-aware fusion model achieved considerably better recognition performance than the uncertainty-unaware model-agnostic fusion baselines. \ch{In recognising dimensional emotions, COLD fusion demonstrated  $\sim$6\% relative improvement over the best-performing fusion baseline, and in the case of categorical emotion recognition it achieved $\sim$8.2\% relative improvement over the existing state-of-the-art model.} Validating our main hypothesis, extensive ablation studies (see Appendix.~\ref{app:ablation}) showed that it is important to apply both calibration and ordinality constraints for improving the emotion recognition results of uncertainty-aware fusion models. Furthermore, our method demonstrated noticeable improvements in terms of predictive uncertainty calibration errors of the emotion recognition models. It is important to note that our proposed calibration and ordinal ranking constraints can be easily applied to general model-fusion methods as well by quantifying the model-wise predictive uncertainty values of emotion labels. Future work can consider evaluating the COLD fusion approach on other complex multimodal learning tasks such as audiovisual speech recognition in noisy conditions~\cite{stewart2013robust} and humour detection~\cite{christ2022muse,christ2023muse}, etc.

In summary, this work showed the importance of uncertainty modelling for the dynamic integration of emotional expression cues from multimodal signals. We believe that uncertainty-aware information fusion is fundamental to reliably recognise apparent emotional states in naturalistic conditions. We hope that the results we demonstrated in this work may help in generating more interest in embracing uncertainty in multimodal affective computing.

\ifCLASSOPTIONcompsoc
  \section*{Acknowledgments}
\else
  \section*{Acknowledgment}
\fi

The work of Mani Tellamekala was funded  by  the  Engineering and  Physical Science  Research  Council project  (2159382)  and  Unilever  U.K. Ltd, and Michel Valstar's work was supported by  the Nottingham Biomedical Research Centre.

\ifCLASSOPTIONcaptionsoff
  \newpage
\fi

\bibliographystyle{IEEEtran}
\bibliography{egbib}

\begin{IEEEbiography}[{\includegraphics[width=1.1in,height=1.1in,clip,keepaspectratio]{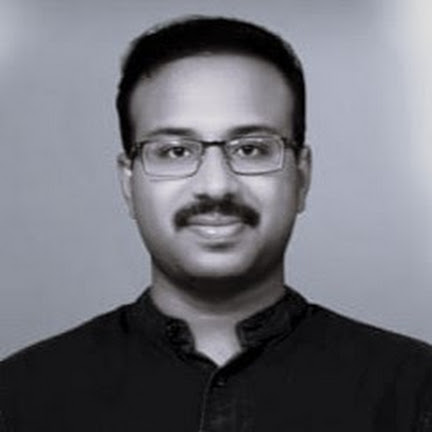}}]%
{Mani Kumar Tellamekala}  is currently a PhD student at the Computer Vision Laboratory, University of Nottingham, UK. He received his Bachelor's degree in Electronics and Communication Engineering from IIIT-RGUKT RK Valley, India. His research focuses on uncertainty-aware temporal learning and self-supervised representation learning, and their applications to affect recognition from face and voice data.  
\end{IEEEbiography}


\begin{IEEEbiography}[{\includegraphics[width=1.1in,height=1.1in,clip,keepaspectratio]{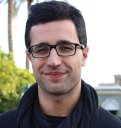}}]%
{Shahin Amiriparian} received his Doctoral degree with the highest honours (summa cum laude) at the Technical University of Munich, Germany in 2019. Currently, he is a postdoctoral researcher at the Chair of Embedded Intelligence for Health Care and Wellbeing, University of Augsburg, Germany. His main research focus is deep learning, unsupervised representation learning, and transfer learning for machine perception, affective computing, and audio understanding. 
\end{IEEEbiography}


\begin{IEEEbiography}[{\includegraphics[width=1.3in,height=1.3in,clip,keepaspectratio]{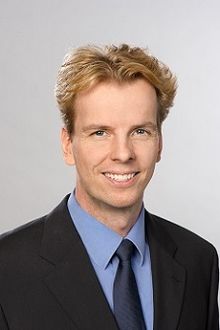}}]%
{Björn W. Schuller} (M’06, SM’15, Fellow’18) received the Diploma in 1999, the Doctoral degree in 2006, and the Habilitation and Adjunct Teaching Professorship in the subject area of signal processing and machine intelligence in 2012, all in electrical engineering and information technology from the Technical University of Munich, Germany. He is Professor of Artificial Intelligence in the Department of Computing, Imperial College London, U.K., a Full Professor and head of the Chair of Embedded Intelligence for Health Care and Wellbeing at the University of Augsburg, Germany. He (co-)authored five books and more than 1200 publications in peer-reviewed books, journals, and conference proceedings leading to more than 46000 citations (h-index = 97).
\end{IEEEbiography}


\begin{IEEEbiography}[{\includegraphics[width=1.1in,height=1.1in,clip,keepaspectratio]{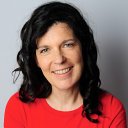}}]
{Elisabeth André} is a full professor of computer science and founding chair of human-centered artificial intelligence at Augsburg University, Augsburg, Germany. Her work has won many awards including the Gottfried Wilhelm Leibnitz Prize 2021, with 2.5 Mio € the highest endowed German research award. In 2010, she was elected a member of the Academy of Europe and the German Academy of Sciences Leopoldina. In 2017, she was elected to the CHI Academy, an honorary group of leaders in the field of human–computer interaction. To honor her achievements in bringing artificial intelligence techniques to human–computer interaction, she was awarded a EurAI fellowship in 2013. From 2019 - 2022, she has been serving as the editor-in-chief of IEEE Transactions on Affective Computing. 
\end{IEEEbiography}


\begin{IEEEbiography}[{\includegraphics[width=1in,height=1in,clip,keepaspectratio]{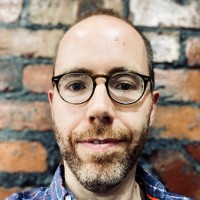}}]%
{Timo Giesbrecht} is a Consumer Science Leader in Beauty \& Personal Care Science \& Technology at Unilever R\&D Port Sunlight. He holds a PhD from Maastricht University. In his work, he employs a wide array of measures and paradigms such as sensory, neuroimaging, psychophysiological measures, and behavioural tasks in collaboration with leading academics around the world. He published more than 95 scientific papers and book chapters and his Hirsch Index is 35. For his research, he has received funding from Dutch and UK Research Councils. Before joining Unilever, Timo Giesbrecht worked at Maastricht University, the Netherlands and Mount Sinai School of Medicine, New York.
\end{IEEEbiography}


\begin{IEEEbiography}[{\includegraphics[width=1.1in,height=1.1in,clip,keepaspectratio]{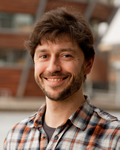}}]%
{Michel Valstar} is Professor in Automatic Understanding of Human Behaviour with the University of Nottingham, School of Computer Science. He is a member of both the Computer Vision Lab and the Mixed Reality Lab. Automatic Human Behaviour Understanding encompasses Machine Learning analysis of data from observable human behaviour and a good idea of how and why people behave in this world. Valstar was a visiting researcher at the Affective Computing group at the Media Lab, MIT, and a research associate with the iBUG group, Department of Computing, Imperial College London. His expertise is facial expression recognition, in particular the analysis of FACS Action Units. He recently proposed a new field of research called `Behaviomedics’, which applies affective computing and Social Signal Processing to the field of medicine to help diagnose, monitor, and treat medical conditions that alter expressive behaviour such as depression.  
\end{IEEEbiography}

\newpage
\appendices

\section{Comparison with a Multimodal Transformer~\cite{tsai2019multimodal}}
\label{app:transformer}

\noindent In addition to the standard fusion baselines, we implemented a multimodal Transformer model based on pair-wise crossmodal self-attention fusion proposed in Tsai et al.~\cite{tsai2019multimodal}. It is worth noting that the crossmodal self-attention fusion aims to cope with the problem of temporal misalignment between different modalities during fusion, similar to the temporal context fusion model we evaluated in this work.  We implemented an audio-visual version of this multimodal Transformer method by tailoring its original network architecture designed for the text, audio, and visual modalities\footnote{\url{https://github.com/yaohungt/Multimodal-Transformer}}. We used a 3-layer self-attention network with 16 heads followed by an FC output layer to implement this multimodal Transformer. As shown in~\Cref{tab:comp_attn}, regression results on the AVEC 2019 CES corpus show that the COLD fusion clearly outperformed the Transformer baseline, particularly in arousal prediction, by a large margin.  
\begin{table}[h!]
    \centering
    \renewcommand{\arraystretch}{1.3} 
    \begin{tabular}{l r r r}
    \toprule
    & \textbf{Valence} & \textbf{Arousal} & \textbf{Avg.} \\
    \textbf{Model} & \textbf{CCC $\uparrow$} & \textbf{CCC $\uparrow$} & \textbf{CCC $\uparrow$} \\
    \midrule
    Multimodal AV Transformer~\cite{tsai2019multimodal}$^\dagger$ & 0.602 & 0.619 & 0.610 \\
    Proposed AV COLD Fusion & \textbf{0.611} & \textbf{0.661} & \textbf{0.636} \\
    \bottomrule 
    \end{tabular}
    \caption{Comparison with a pair-wise crossmodal self-attention based multimodal Transformer~\cite{tsai2019multimodal} ($^\dagger$ indicates in-house implementation for AV fusion): Regression results on the AVEC 2019 CES validation set.}
    \label{tab:comp_attn}
\end{table}

\section{Ablation Studies}
\label{app:ablation}

\noindent \Cref{tab:ablation} presents the ablation results that quantify the contributions of calibration, ordinality, and variance regularisation constraints to the performance gains achieved by COLD fusion. By individually nullifying the four optimisation hyperparameters of the COLD training objective (see~\Cref{eq:loss_main}), we measure the emotion regression performance on the AVEC 2019 validation set. Compared to the fully constrained COLD fusion model, different partially constrained and fully unconstrained models listed in~\Cref{tab:ablation}, achieve considerably lower CCC scores. Most importantly, we observe that discarding the variance regularisation constraint results in more performance degradation than the remaining constraints. This observation indicates the importance of preventing the variance collapse problem by using the variance regularisation term, in line with the results reported in prior works~\cite{sanchez2021affective,chang2020data}. 

\begin{table}[h!]
    \centering
    \begingroup
    \renewcommand{\arraystretch}{1.3}        
    \begin{tabular}{l r r}
    \toprule
    \textbf{Model} & \textbf{Valence} & \textbf{Arousal} \\
    & CCC $\uparrow$ & CCC $\uparrow$ \\
    \midrule
    \textit{With All three constraints}  & \textbf{0.605} & \textbf{0.661} \\    
    ($\lambda_{CV}=\lambda_{CA}=\lambda_C{_{AV}}=1e-3$, $\lambda_R=1e-4$) & & \\
    \textit{Without Intramodal constraints} ($\lambda_{CV}=\lambda_{CA=0}$) & 0.573 & 0.615 \\
    \textit{Without Crossmodal constraint} ($\lambda_C{_{AV}}$) & 0.580 & 0.609 \\
    \textit{Without Regularisation constraint} ($\lambda_R=0$) & 0.541 & 0.595 \\
    \textit{Without Any constraints} & 0.517 & 0.578 \\
    ($\lambda_{CV}=\lambda_{CA}=\lambda_C{_{AV}}=0$, $\lambda_R=0$) & & \\
    \bottomrule
    \end{tabular}
    \caption{Ablation experiments on the proposed loss function (\Cref{eq:L_CO}): Analysing the impact of different loss components in the COLD Fusion on the AVEC 2019 CES validation set (CCC-Concordance Correlation Coefficient).}
    \label{tab:ablation}
    \endgroup
\end{table}

 \section{Statistical Significance Analysis}
 \label{app:stat_sig}
 
\noindent As shown in~\Cref{tab:stat_sig}, we conducted paired $t$-test on the validation set of the AVEC 2019 corpus, to verify the statistical significance of COLD fusion's performance improvements over the unimodal and remaining multimodal baselines. In line with the trends in regression, performance reported in~\Cref{tab:avec19_reg}, p-values of the student $t$-test indicate that the improvements achieved by the COLD fusion models are statistically quite significant compared to the baseline models.   \\  

\begin{table}[h!]
    \centering
    \begingroup
    \renewcommand{\arraystretch}{1.3}        
    \begin{tabular}{l r r}
    \toprule
    Model Pair & Valence & Arousal \\
    & p-Value $\downarrow$ & p-Value $\downarrow$ \\
    \midrule
    (Aud-branch, AV COLD Fusion) & 6.4e-34 & 9.7e-23 \\
    (Vis-branch, AV COLD Fusion) & 1.9e-15 & 3.3e-16 \\
    (AV Feature Fusion, AV COLD Fusion) & 3.7e-14 & 1.1e-18\\
    (AV Prediction Fusion, AV COLD Fusion) & 5.5e-9 & 7.3e-3\\
    (AV Context Fusion, AV COLD Fusion) & 1.2e-4 & 2.0e-3\\
    (AV COLD Fusion, AV COLD Fusion) & 1.0e-0 & 1.0e-0 \\
    \bottomrule
    \end{tabular}
    \caption{Statistical significance testing ($p<$0.01): \textit{Regression} $t$-test results on the {AVEC 2019 CES validation set}.}
    \label{tab:stat_sig}
    \endgroup
\end{table}

\section{Reliability Diagrams: Uncertainty Calibration Performance Evaluation}
\label{app:reliability}
\begin{figure}
    \includegraphics[width=1.0\linewidth]{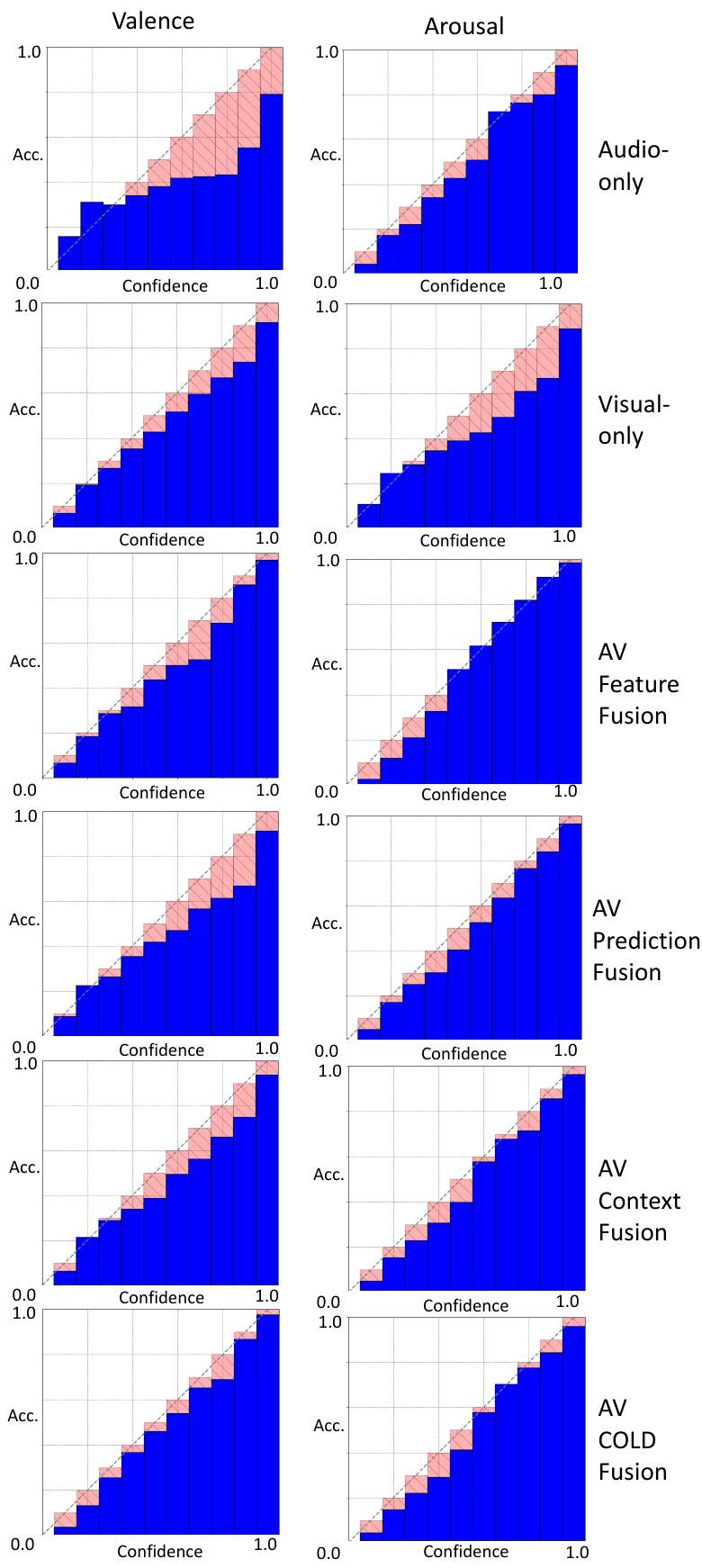}
    \caption{Reliability plots of unimodal and multimodal classification models evaluated on the AVEC 2019 validation set. A perfectly calibrated model appears as a perfect right angled triangle, as marked by the diagonal lines and the red bars.}
    \label{fig:reliab_plots}
\end{figure}

\noindent Reliability diagrams visually illustrate the uncertainty calibration performance of a model's predictions. As~\Cref{fig:reliab_plots} shows, when a model is perfectly calibrated, its confidence score vs the accuracy score histogram looks like a perfect right-angled triangle. The more the deviations are from the diagonal lines in them, the higher their ECE values are. Note that ECE is a scalar summary statistic of a reliability diagram, which computes the weighted average of such deviations over all the intervals in the reliability diagram. Though the ECE values reported for both the AVEC 2019 corpus (\Cref{tab:avec19_calib}) and IEMOCAP (\Cref{tab:iemocap_calib}) already validate the improved calibration results with COLD fusion. Here, as an example, 
 in~\Cref{fig:reliab_plots} we compare the reliability plots of different models evaluated on the AVEC validation set.  In~\Cref{fig:reliab_plots}, we can see that compared to the unimodal cases and other fusion baselines, the COLD fusion reliability plot looks much closer to a perfect right-angled triangle. Among all the reliability plots illustrated, we observe that the audio branch for valence has the highest calibration error. This observation is in line with the poor performance achieved by the audio modality in terms of the valence prediction error (see~\Cref{tab:avec19_reg} and~\Cref{tab:avec19_classif}) on the AVEC 2019 corpus.    \\

\section{Application of COLD Fusion to Multimodal Sentiment Intensity Prediction}
\label{app:mosei}

\noindent \textbf{Experimental Setup.} In this evaluation our  objective is to analyse the sentiment recognition performance of the proposed uncertainty-aware temporal context modelling approach, by demonstrating its application to multimodal fusion on a large-scale in-the-wild audiovisual dataset. Considering that the simple recurrent temporal model is used in the COLD fusion mechanism implemented in this work, we also show that the proposed uncertainty-aware fusion step can be easily integrated into the existing multimodal fusion techniques. We demonstrate this by modifying the standard modality-wise LSTM / GRU layer output layers in an existing multimodal fusion network, and by adding additional loss terms specific to the COLD fusion as described in~\Cref{eq:loss_main}. Based on these modifications, we added the COLD fusion step in one of the existing benchmark fusion models on CMU-MOSEI, Modality-Invariant and -Specific Representations (MISA)~\cite{hazarika2020misa}. MISA is a simple multimodal fusion approach in which utterance-level embeddings of the audio, visual and language modalities are projected into modality-invariant and modality-specific subspaces using different encoder modules. To extract the unimodal utterance embeddings in MISA, stacked LSTM modules are applied to each modality separately, similar to the temporal context extraction step followed in the COLD fusion model.

    \begin{table}[h]
        \centering
        \begingroup
        \renewcommand{\arraystretch}{1.3}    
        \begin{tabular}{l  r r r  r r }
        \toprule
              \textbf{Model} &  \textbf{MAE $\downarrow$} & \textbf{Corr $\uparrow$} & \textbf{Acc-7 $\uparrow$} &  \textbf{Acc-2 $\uparrow$} & \textbf{F1 $\uparrow$} \\
            \hline
        Graph-MFN~\cite{zadeh2018multimodal} & 0.710 & 0.540 & 45.0 & 76.9 & 77.0 \\
        RAVEN~\cite{wang2019words} & 0.614 & 0.662 & 50.0 & 79.1 & 79.5 \\
        MCTN~\cite{pham2019found} & 0.609 & 0.670 & 49.6 & 79.8 & 80.6 \\
        LMF~\cite{liu2018efficient} & 0.623 & 0.677 & 48.0 & 82.0 & 82.1 \\
        TFN~\cite{zadeh2017tensor} & 0.593 & 0.700 & 50.2 & 82.5 & 82.1 \\
        MulT~\cite{tsai2019multimodal} & 0.570 & 0.758 & 51.1 & 84.5 & 84.5 \\
        MFM~\cite{tsai2018learning} & 0.568 & 0.717 & 51.3 & 84.4 & 84.3 \\
        ICCN~\cite{sun2020learning} & 0.565 & 0.713 & 51.6 & 84.2 & 84.2 \\
        \midrule
        MISA~\cite{hazarika2020misa}$^*$ & 0.555 & 0.756 & 52.2 & 85.5 & 85.3 \\
        MISA~\cite{hazarika2020misa}$^\dagger$ & 0.557 & 0.748 & 51.7 & 84.9 & 84.8 \\
        COLD Fusion & 0.549 & 0.752 & 52.1 & 85.2 & 85.0   \\
        \textit{MISA~\cite{hazarika2020misa}$^\dagger$ with} & \\
        \textit{COLD Fusion} & \textbf{0.548} & \textbf{0.760} & \textbf{52.4} & \textbf{85.8} & \textbf{85.5} \\
        \bottomrule
        \end{tabular}
        \caption{Multimodal sentiment regression and classification results on the test set of \textbf{CMU-MOSEI} dataset ($*$ denotes the originally reported results and $\dagger$ denotes the results that we could reproduce using the publicly available implementation) }
        \label{tab:mosei_classif_exp}
        \endgroup
    \end{table}

We followed the same audio, visual and textual feature extraction steps implemented in the original MISA model. Here we applied the COLD fusion to only audio and visual features, as the text embeddings are prepared using a pre-trained BERT model~\cite{devlin2018bert}. Building on the main idea of MISA, we replaced the modality-wise (audio and visual) recurrent temporal modules in the MISA implementation with the modified GRU layers that output mean and variance vectors as in the COLD fusion model. Before learning the modality-invariant and modality-specific features, the temporal context embeddings of audio and visual modalities are multiplied with their corresponding COLD fusion weight vectors (see~\Cref{eq:fusion_weights}). For the purpose of computing the COLD fusion loss (\Cref{eq:loss_main}), we additionally included modality-wise sentiment prediction modules, composed of a 2-layer fully connected network on top of the GRU layer outputs. Note that all the remaining experimental setup details of this modified MISA model that we used here for the COLD fusion evaluation are the same as in the publicly available implementation of MISA\footnote{MISA - \url{https://github.com/declare-lab/MISA}}. \\

\noindent \textbf{Evaluation Criteria.} For analysing the sentiment intensity prediction performance in regression tasks, we used mean absolute error (MAE) and Pearson's correlation coefficient (Corr) as evaluation metrics. Similar to the existing benchmarks, we also included the classification results in terms of 7-class accuracy (Acc-7), 2-class accuracy of positive and negative classification task (Acc-2) and its corresponding F1 score values.  \\

\noindent \textbf{Results and Analysis.} In~\Cref{tab:mosei_classif_exp} we present the sentiment intensity regression and classification results of the MISA model with and without COLD fusion on the test set of CMU-MOSEI, in comparison with the following fusion methods evaluated on multimodal sentiment analysis tasks: 

\begin{itemize}
    \item A graph-based fusion model (Graph-MFN~\cite{zadeh2018multimodal})
    \item Attention and Transformer-based fusion models (RAVEN~\cite{wang2019words} and MulT~\cite{tsai2019multimodal})
    \item Fusion models based on subspace learning  (MCTN~\cite{pham2019found} and MFM~\cite{tsai2018learning})
    \item A tensor-based fusion model (TFN~\cite{zadeh2017tensor}) and its low-rank variant  (LMF~\cite{liu2018efficient})
    \item A canonical correlation-based fusion network (ICCN~\cite{sun2020learning}) 
\end{itemize}

Compared to the above-listed fusion models, the proposed COLD fusion mechanism integrated with the MISA model shows consistently better sentiment intensity prediction results in terms of both classification and regression metrics. The overall performance improvements achieved by the MISA + COLD fusion model validate the main hypothesis of this work that the uncertainty-aware temporal fusion improves the model's predictive performance. The performance difference between the MISA models with and without COLD fusion, indicates that the COLD fusion mechanism can complement the existing uncertainty-unaware temporal fusion models, at the cost of requiring only minimal changes to the canonical temporal model's architecture and the training loss function. 

\end{document}